\pdfoutput=1

\documentclass[11pt]{article}

\usepackage{acl}

\usepackage{times}
\usepackage{latexsym}

\usepackage[T1]{fontenc}

\usepackage[utf8]{inputenc}

\usepackage{microtype}

\usepackage{algorithm}
\usepackage{algorithmic}
\usepackage{graphicx}

\usepackage{makecell}
\newcommand{\PreserveBackslash}[1]{\let\temp=\\#1\let\\=\temp}
\newcolumntype{C}[1]{>{\PreserveBackslash\centering}p{#1}}
\newcolumntype{R}[1]{>{\PreserveBackslash\raggedleft}p{#1}}
\newcolumntype{L}[1]{>{\PreserveBackslash\raggedright}p{#1}}
\usepackage{subfig}
\usepackage{multirow}
\usepackage{amssymb}
\usepackage{amsfonts}
\usepackage{amsmath}
\usepackage{booktabs}
\usepackage{CJKutf8}

\newcommand{\redbold}[1]{\textcolor{red}{\textbf{#1}}}
\newcommand{\blueit}[1]{\textcolor{blue}{\textit{#1}}}

\title{CVLUE: A New Benchmark Dataset \\ for Chinese Vision-Language Understanding Evaluation}

\author{Yuxuan Wang$^{1}$, Yijun Liu$^{2}$, Fei Yu$^{1}$, Chen Huang$^{1}$, Kexin Li$^{1}$, Zhiguo Wan$^{1}$, Wanxiang Che$^{2}$ \\
$^{1}$Zhejiang Lab, Hangzhou, 311121 \\
$^{2}$Harbin Institute of Technology, Harbin, 150001 \\
\{yxwang, yufei, huangc, likx, wanzhiguo\}@zhejianglab.com \\
\{yijunliu, car\}@ir.hit.edu.cn }

\begin{document}
\maketitle
\begin{abstract}

Despite the rapid development of Chinese vision-language models (VLMs), most existing Chinese vision-language (VL) datasets are constructed on Western-centric images from existing English VL datasets. 
The cultural bias in the images makes these datasets unsuitable for evaluating VLMs in Chinese culture. 
To remedy this issue, we present a new Chinese Vision-Language Understanding Evaluation (CVLUE) benchmark dataset, where the selection of object categories and images is entirely driven by Chinese native speakers, ensuring that the source images are representative of Chinese culture. 
The benchmark contains four distinct VL tasks ranging from image-text retrieval to visual question answering, visual grounding and visual dialogue.
We present a detailed statistical analysis of CVLUE and provide a baseline performance analysis with several open-source multilingual VLMs on CVLUE and its English counterparts to reveal their performance gap between English and Chinese.
Our in-depth category-level analysis reveals a lack of Chinese cultural knowledge in existing VLMs. 
We also find that fine-tuning on Chinese culture-related VL datasets effectively enhances VLMs' understanding of Chinese culture. 
\footnote {Our benchmark and the evaluation codes are available on \url{https://github.com/WangYuxuan93/CVLUE}.}

\end{abstract}

\section{Introduction}

Over the last few years, vision-language pre-training (VLP), as a thriving field, has been drawing extensive attention \cite{lu-etal-2019-vilbert,chen-etal-2020-uniter,cho-etal-2021-unifying,li-etal-2021-align}, leading to significant performance boosts across many VL tasks. 
It cannot be neglected that the abundance of VL datasets covering various distinct VL tasks \cite{young-etal-2014-flickr30k,kazemzadeh-etal-2014-referitgame,antol-etal-2015-vqa,chen-etal-2015-cococaption,mao-etal-2016-refcocog,das-etal-2017-visual,goyal-etal-2017-vqa2} plays an essential role in the rapid evolvement of VLMs. 
However, most of the existing VL datasets are in English. 
A majority of these datasets, such as NLVR2~\cite{suhr-etal-2019-nlvr2} and MS-COCO~\cite{lin-etal-2014-mscoco}, are built on top of a hierarchy of concepts selected from English WordNet~\cite{miller-1992-wordnet}, resulting in source images with a North American or Western European bias \cite{liu-etal-2021-visually}. 
Beyond the English language and Western cultures where these datasets were created, evidence suggests that both the origin \cite{devries-etal-2019-does} and content \cite{stock-etal-2018-convnets} of such data are skewed. 

\begin{table}[t]
	\centering
	\small
	\begin{tabular}{l|C{0.3cm}cccccc}
		\hline
		Ben. & Lan. & ITR & VQA & VG & VD & VR & IG \\
		\hline
		VLUE & En. & \checkmark & \checkmark & \checkmark &  & \checkmark & \\
        CLiMB & En. &  & \checkmark &  &  & \checkmark & \\
        \hline
        MUGE &  Ch. & \checkmark & & & & & \checkmark \\
        Zero  & Ch. & \checkmark & & & &\\
        \textbf{CVLUE} & Ch. & \checkmark & \checkmark & \checkmark & \checkmark & & \\
		\hline
	\end{tabular}
	\caption{Tasks included in CVLUE, VLUE, CLiMB, MUGE and Zero. Ben. and Lan. denote Benchmark and Language, respectively. En. and Ch. stand for English and Chinese respectively. }
	\label{tbl:benchmark-tasks}
\end{table}

Recently, the community has begun to recognize the importance of cultural differences in large language models (LLMs). 
Some work has explored the varied performance of LLMs across different cultural contexts \cite{wang-etal-2023-countries,li-etal-2024-culture}, while other efforts have focused on creating culturally relevant LLM benchmarks \cite{zhao-etal-2024-world, rao-etal-2024-normad}. 
Additionally, there is a small body of work investigating cultural awareness in VLMs \cite{burdalassen-etal-2024-culturally} and developing multicultural visual question answering \cite{romero-etal-2024-cvqa} and visual language reasoning \cite{liu-etal-2021-visually} datasets. 
However, these datasets often prioritize coverage of different cultures, with limited task categories and data volumes specific to Chinese culture.


\begin{figure*}[htbp]
  \centering
  \subfloat[Image Text Retrieval]
  {\includegraphics[width=0.48\textwidth]{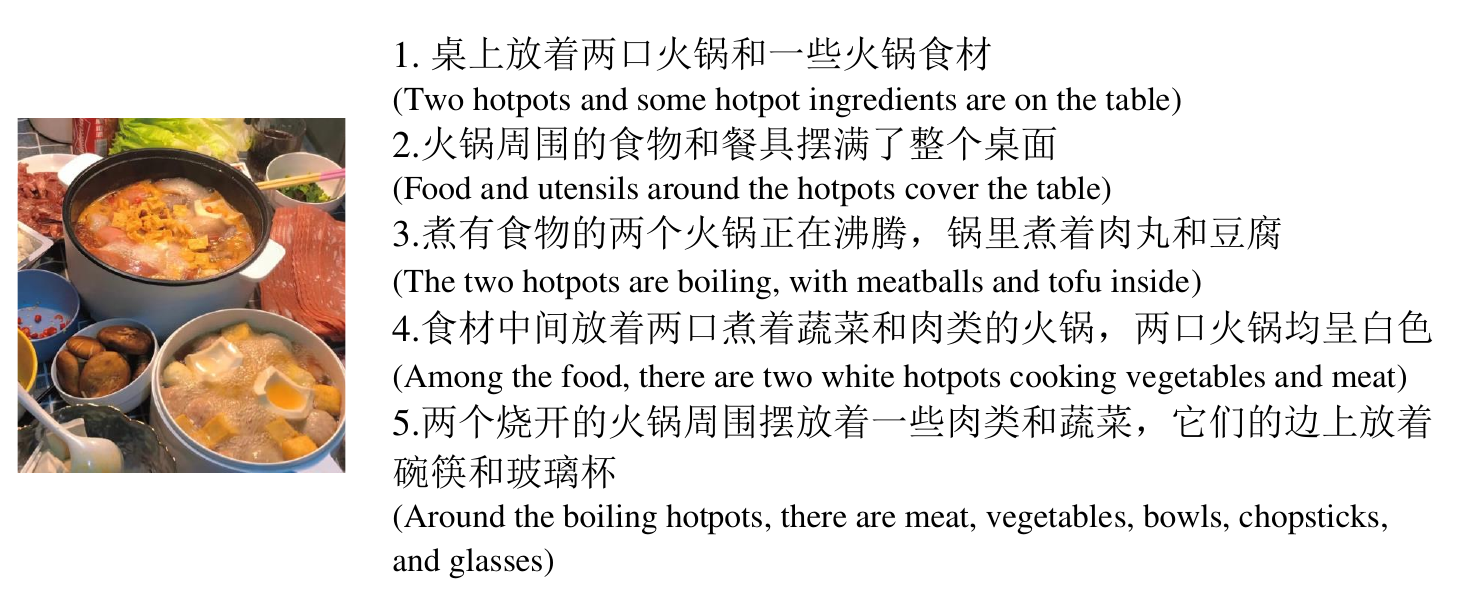}\label{fig:ic-example}}
  \quad     
  \subfloat[Visual Question Answering]
  {\includegraphics[width=0.48\textwidth]{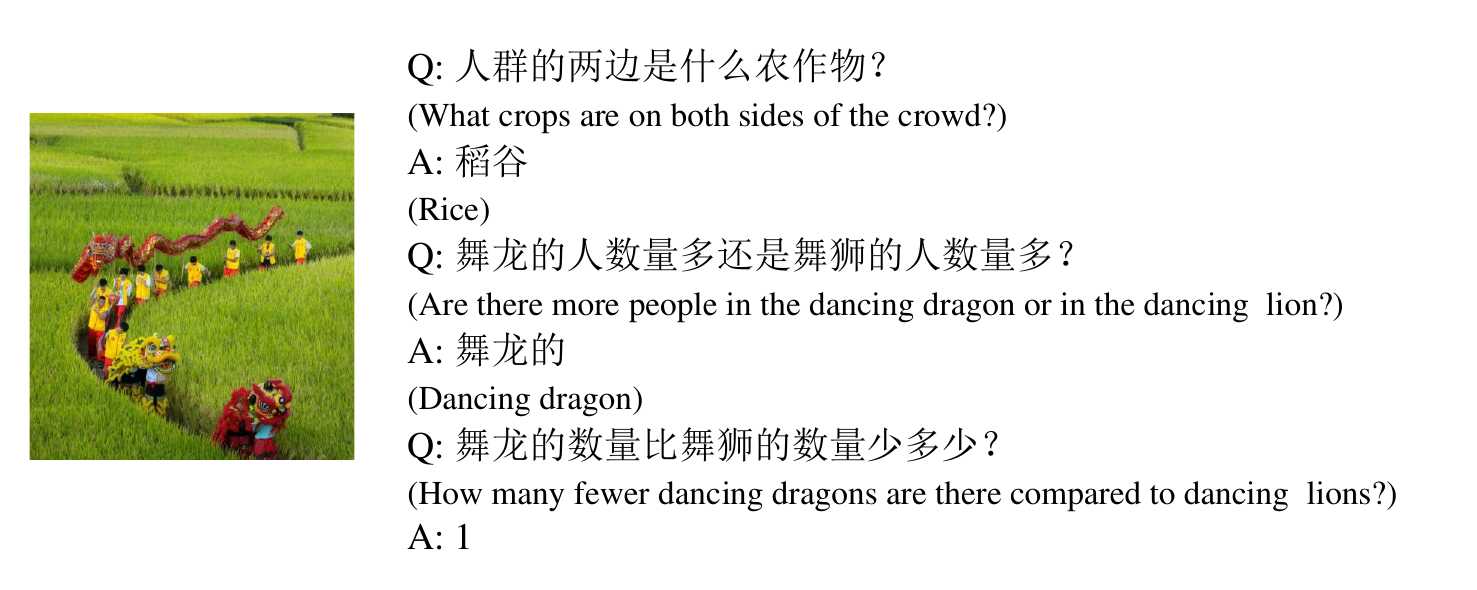}\label{fig:vqa-example}}
  \quad 
  \subfloat[Visual Grounding]
  {\includegraphics[width=0.48\textwidth]{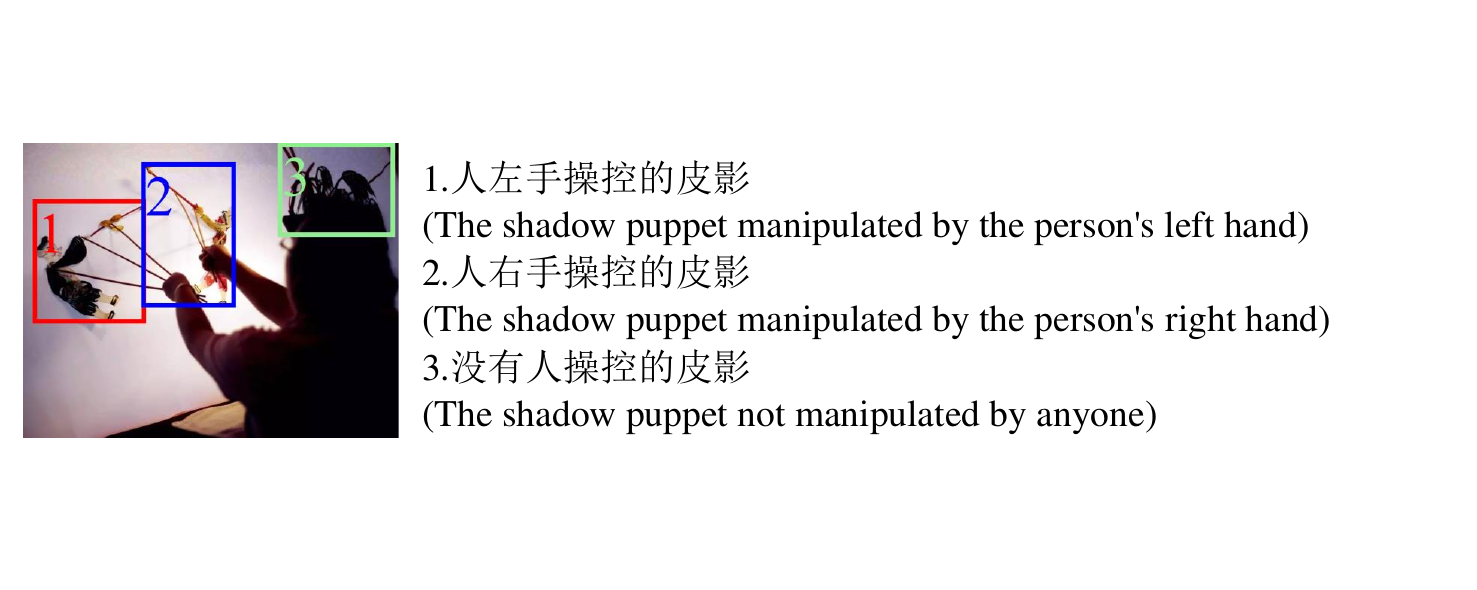}\label{fig:vg-example}}
  \quad 
  \subfloat[Visual Dialogue]
  {\includegraphics[width=0.48\textwidth]{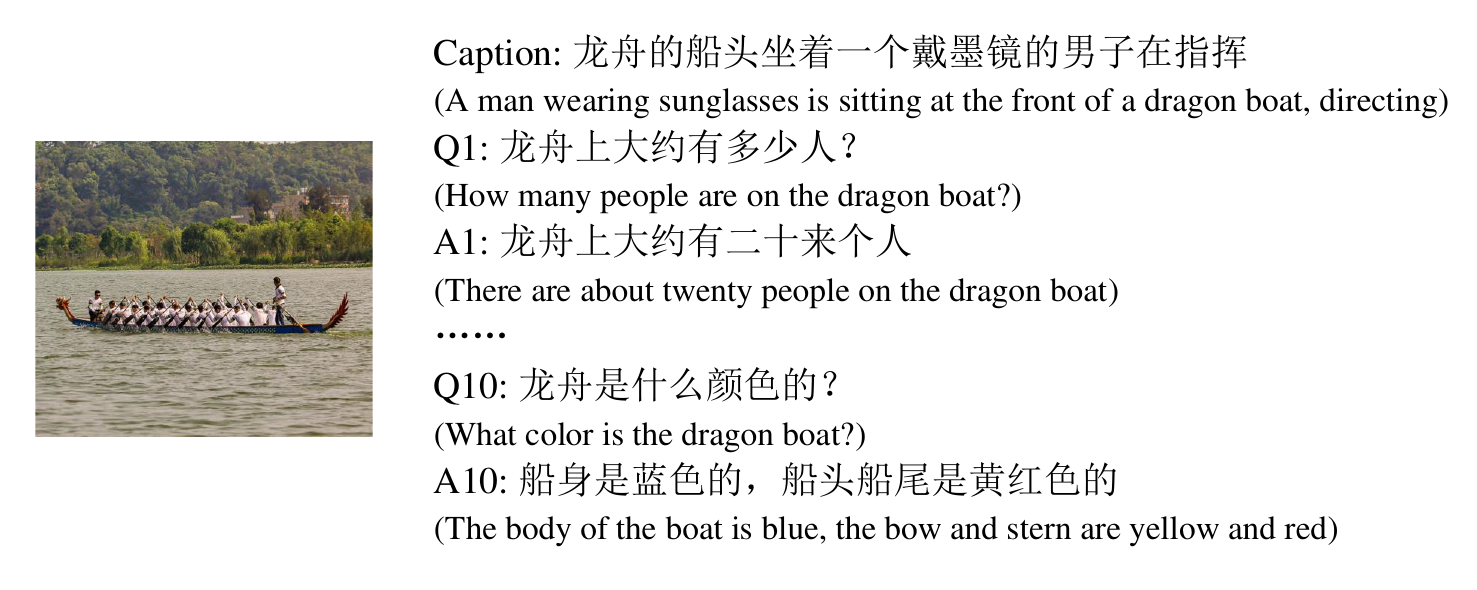}\label{fig:vd-example}}
  \quad 
  \caption{Examples of the images and their annotation for the four tasks in CVLUE.}\label{fig:example}
\end{figure*}

In this work, we focus on the \textit{evaluation of VLMs in Chinese culture, meaning that not only are the texts in Chinese but, more importantly, the images are representative of Chinese culture}. 
Over the last two years, a significant number of multimodal datasets for Chinese VLM pre-training have been presented~\cite{zhan-etal-2021-product1m,lin-etal-2021-m6,gu-etal-2022-wukong,liu-etal-2022-taisu}. 
However, the development of the benchmark dataset for Chinese VLM evaluation is lagging behind. 
Many existing Chinese VL datasets exploit images from English VL datasets containing the abovementioned bias. 

Some of them, such as Flickr30K-CN~\cite{lan-etal-2017-flickr30k-cn}, were constructed by translating texts in English VL datasets into Chinese. 
Others, such as FM-IQA~\cite{gao-etal-2015-fm-iqa}, Flickr8K-CN~\cite{li-etal-2016-flickr8k-cn} and COCO-CN~\cite{li-etal-2019-coco-cn}, were constructed by re-annotating images from English VL datasets in Chinese. 
Recently, several new datasets have been presented, whose images were collected from image search engines with Chinese queries. 
However, they are limited to single types of tasks like visual question answering \cite{wang-etal-2022-chiqa} or image-text retrieval \cite{xie-etal-2022-zero-r2d2}.

Chinese is linguistically distinct from English and many other languages, whose speakers comprise one-fourth of the world's population.
This necessitates a benchmark dataset specifically designed for Chinese vision-language understanding (VLU). 
To remedy this issue, we present CVLUE, a new Chinese VL benchmark dataset.
We start by selecting categories representative of Chinese culture and manually collect all the images from the Chinese Internet, ensuring that \textit{the source images are commonly seen or representative in the Chinese-speaking population}. 
The comparison between CVLUE and existing VL benchmark datasets is shown in Table~\ref{tbl:benchmark-tasks}.\footnote{We only compare with benchmarks containing at least two subtasks here.}
The visual reasoning (VR) task is included in the two English benchmark datasets VLUE~\cite{zhou-etal-2022-vlue} and CLiMB~\cite{srinivasan-etal-2022-climb} but not included in any of the Chinese ones. 
The image generation (IG) task is only included by MUGE\footnote{\url{https://tianchi.aliyun.com/muge}}, which mainly contains simple iconic images collected from e-commerce platforms and encyclopedias. 
On the contrary, images in our benchmark were mostly non-iconic ones. 
The other Chinese dataset Zero~\cite{xie-etal-2022-zero-r2d2} only focuses on image-text matching and retrieval and comprises five subtasks of a similar type. 
Our benchmark, by contrast, contains four distinct VL tasks: image-text retrieval (ITR), visual question answering (VQA), visual grounding (VG) and visual dialogue (VD), which evaluate VLMs in Chinese culture from multiple aspects. 
Examples of the images and annotation for the four tasks are shown in Figure~\ref{fig:example}. 
See Appendix~\ref{sec:appendix-cvlue-example} for more. 



We benchmark several popular open-source multilingual VLMs on CVLUE and established English VL datasets to assess their visual language understanding (VLU) capabilities in both Chinese and English. 
Furthermore, our in-depth analysis reveals the lack of Chinese culture-related knowledge in existing VLMs.
We believe this dataset offers a fair and convenient platform for evaluating VLMs in the context of Chinese culture. 


\section{Related Work}
\label{sec:related_work}
Over the last decade, English VL datasets have experienced rapid development, starting from the most fundamental task of image captioning. 
Following the popular MS-COCO~\cite{lin-etal-2014-mscoco} and Flickr30K~\cite{young-etal-2014-flickr30k} datasets, a significant number of VL datasets covering various tasks of visual question answering~\cite{antol-etal-2015-vqa,goyal-etal-2017-vqa2}, visual grounding~\cite{kazemzadeh-etal-2014-referitgame,mao-etal-2016-refcocog}, visual entailment~\cite{xie-etal-2019-visual-entailment}, visual dialogue~\cite{das-etal-2017-visual} and etc. have emerged.
Recently, an increasing number of English VL benchmarks aiming at different goals have been proposed~\cite{parcalabescu-etal-2022-valse,zhou-etal-2022-vlue,zheng-etal-2022-vlmbench,srinivasan-etal-2022-climb}, which significantly facilitates the evaluation and comparison of VLMs in English. 

Beyond the VL datasets in English, MS-COCO was extended with captions translated to or newly written in German and French~\cite{rajendran-etal-2016-bridge}, Japanese~\cite{yoshikawa-etal-2017-stair} and Chinese~\cite{li-etal-2019-coco-cn}. 
All these datasets exploit images crowdsourced from North America and Western Europe. 
Researches suggest that they suffer from cultural bias, which may lead to essential limitations for the application in many languages and cultures \cite{stock-etal-2018-convnets,devries-etal-2019-does,liu-etal-2021-visually}. 
In recent years, the community has begun to notice the performance differences of existing VLMs in different cultural applications~\cite{burdalassen-etal-2024-culturally} and has started to develop multicultural visual question answering \cite{romero-etal-2024-cvqa} and visual language reasoning~\cite{liu-etal-2021-visually} datasets.
However, these datasets focus on broad cultural coverage, resulting in limited task types and data volume for Chinese.


Over the last two years, an increasing number of Chinese multimodal datasets in the form of image-text pairs have been presented \cite{lin-etal-2021-m6,gu-etal-2022-wukong,liu-etal-2022-taisu}, which has dramatically promoted the evolvement of Chinese VLMs. 
However, the development of the benchmark dataset for VLM evaluation in Chinese is lagging behind. 
A great number of existing Chinese VL datasets were constructed by extending English VL datasets with translated~\cite{lan-etal-2017-flickr30k-cn} or newly written~\cite{gao-etal-2015-fm-iqa,li-etal-2016-flickr8k-cn,li-etal-2019-coco-cn} annotation in Chinese. 
\citet{wu-etal-2017-aic} presented a Chinese image captioning dataset AIC-ICC, whose images were newly collected from search engines. 
Recently, two Chinese VQA datasets were introduced, both constructed with newly collected images~\cite{qi-etal-2022-dureader,wang-etal-2022-chiqa}. 
However, these datasets are limited to single types of tasks and thus insufficient for the comprehensive evaluation of VLMs. 

Due to the abundance of English VL datasets, recent English VL benchmarks were mainly constructed using existing datasets. 
However, given the situation of existing Chinese VL datasets, building a benchmark specifically for Chinese is much more challenging.
Recently, \citet{xie-etal-2022-zero-r2d2} introduced a new Chinese VL dataset Zero covering five subtasks. 
However, all of them involve image-text retrieval/matching and are, therefore, not comprehensive enough to evaluate the general capability of VLMs. 
\citet{liu-etal-2023-mmbench} proposed a bilingual VL benchmark MMBench, which is first annotated in English and then translated to Chinese using GPT-4. 
Interestingly, they also released CCBench\footnote{\url{https://github.com/open-compass/MMBench}}, a 510-example multiple-choice question answering test set with images closely related to Chinese culture. 
While it aligns most closely with the goals of this paper, it has significantly less diversity in task types and annotated data than CVLUE.

\begin{CJK}{UTF8}{gbsn}
\begin{table*}[htbp]
	\centering
	\small
	\begin{tabular}{L{0.10\textwidth}|L{0.84\textwidth}}
		\hline
		\bf Semantic Fields & \bf Categories  \\
		\hline
		\bf Animal & 大熊猫 (panda), \blueit{牛 (cow)}, 鱼 (fish), \blueit{狗 (dog)}, \blueit{马 (horse)}, 鸡 (chicken), \blueit{鼠 (mouse)}, \blueit{鸟 (bird)}, \blueit{人 (human)}, \blueit{猫 (cat)} \\
        \bf Food   & \redbold{火锅 (hot pot)}, 米饭 (rice), 饺子 (dumpling), 面条 (noodles), \redbold{包子 (stuffed bun)} \\
        \bf Beverages & \redbold{奶茶 (bubble tea)}, 可乐 (coke), 牛奶 (milk), 茶 (tea), 粥 (porridge), 酒 (alcohol) \\
        \bf Clothing  & \redbold{汉服 (Hanfu)}, \redbold{唐装 (Tang suit)}, \redbold{旗袍 (cheongsam)}, 西装 (suit), T恤 (T-shirt) \\
        \bf Plant  & 柳树 (willow), 银杏 (ginkgo), 梧桐 (Chinese parasol), 白桦 (birch), 松树 (pine), 菊花 (chrysanthemum), 牡丹 (peony), 兰花 (orchid), 莲 (lotus), 百合 (lily) \\
        \bf Fruit  & 荔枝 (lychee), 山楂 (hawthorn), \blueit{苹果 (apple)}, 哈密瓜 (cantaloupe), 龙眼 (longan) \\ 
        \bf Vegetable & 小白菜 (bok choy), 马铃薯 (potato), 大白菜 (Chinese cabbage), 胡萝卜 (carrot), \blueit{花椰菜 (cauliflower)} \\
        \bf Agriculture & 锄头 (hoe), 犁 (plow), 耙 (harrow), 镰刀 (sickle), \redbold{担杖 (carrying pole)} \\
        \bf Tool   & \blueit{汤勺 (spoon)}, \blueit{碗 (bowl)}, 砧板 (cutting board), 筷子 (chopsticks), 炒锅 (wok), 扇子 (fan), \redbold{菜刀 (Chinese cleaver)}, \redbold{锅铲 (wok spatula)} \\
        \bf Furniture & \blueit{电视 (TV)}, 桌子 (table), \blueit{椅子 (chair)}, \blueit{冰箱 (refrigerator)}, 灶台 (cooking stove) \\
        \bf Sport  & 乒乓球 (Ping-Pong), 篮球 (basketball), 游泳 (swimming), 足球 (football), 跑步 (running) \\
        \bf Celebrations & \redbold{舞狮 (lion dance)}, \redbold{龙舟 (dragon boat)}, 国旗 (national flag), \redbold{月饼 (mooncake)}, 春联 (couplet), 花灯 (lantern) \\
        \bf Education & 铅笔 (pencil), 黑板 (blackboard), \redbold{毛笔 (Chinese brush)}, 粉笔 (chalk), 原子笔 (ballpoint), \blueit{剪刀 (scissors)} \\
        \bf Instruments & \redbold{古筝 (Chinese zither)}, \redbold{二胡 (erhu)}, \redbold{唢呐 (suona)}, 鼓 (drums), \redbold{琵琶 (pipa)} \\
        \bf Arts   & \redbold{毛笔书法 (brush calligraphy)}, \redbold{皮影 (Chinese shadow play)}, \redbold{剪纸 (paper cutting)}, \redbold{兵马俑 (Terracotta Army)}, \redbold{鼎 (ding)}, 陶瓷 (ceramics) \\
		\hline
	\end{tabular}
	\caption{Object categories in CVLUE, where the 15 categories overlapping with MS-COCO are shown in blue italic font, while the 22 categories not in WordNet are shown in red bold font.}
	\label{tbl:category}
\end{table*}
\end{CJK}

\section{CVLUE}


Our dataset consists of four distinct VL tasks that evaluate a model's capability in Chinese VLU from multiple aspects. 
The data splits and evaluation metrics are summarized in Table~\ref{tbl:characteristics}.
In this section, we describe the procedure we devised for image collection and dataset annotation.

\begin{table}[ht]
	\centering
	\small
	\begin{tabular}{l|cccl}
		\hline
		Task & $|$Train$|$ & $|$Valid$|$ & $|$Test$|$ & Metrics \\
		\hline
        ITR & 17,920 & 3,116 & 8,973 & R@k \\
        VQA & 14,362 & 2,571 & 7,169 & Acc \\
        VG  & 10,769 & 1,965 & 5,385 & IoU \\
        VD  & 3,975  & 651   & 2,036 & R@k \\
		\hline
	\end{tabular}
	\caption{Data splits (in terms of image numbers) and evaluation metrics of tasks in CVLUE. R@k denotes the recall in the top k predictions, Acc stands for accuracy, and IoU stands for intersection over union. }
	\label{tbl:characteristics}
\end{table}

\subsection{Selection of Object Categories}
\label{sec:selection-of-object-categories}


We first explain the selection of object categories, which must form a representative set of categories in Chinese daily life and reflect the unique characteristics of Chinese culture. 
The selection process for our dataset was inspired by the Chinese part of MaRVL~\cite{liu-etal-2021-visually}, where five native speakers provided 5-10 specific concepts for 18 semantic fields, ensuring they are commonly seen, representative, physical, and concrete. 
However, since CVLUE is specifically for Chinese, MaRVL's categories are not directly applicable.

Therefore, we first removed categories not strongly related to specific objects with clear boundaries (e.g., Taoism).
We also replaced some categories with more concrete categories that have clearer boundaries (e.g., replacing the Dragon Boat Festival with dragon boat, replacing the Mid-Autumn Festival with moon cake). 
Then, we merged some categories to make sure that all categories occurred frequently enough so that we could collect enough images for each of them (e.g., merging all types of birds into one bird category). 
Besides, we added some categories representative of Chinese culture (e.g., stuffed buns, fans).

Eventually, we selected 92 object categories from 15 semantic fields listed in Table~\ref{tbl:category}. 
The 15 categories overlapping with MS-COCO (e.g., human, dog), shown in blue italic font, can be regarded as having the weakest association with Chinese culture. 
The 22 categories not in English WordNet~\cite{miller-1992-wordnet} (e.g., guzheng, suona), shown in red bold font, are considered to be culturally closest to Chinese. 
The remaining categories have a moderate association.

\subsection{Task Selection}

As introduced in section~\ref{sec:related_work}, there are currently a wide variety of VL tasks. 
Due to budgetary constraints, we focused on the following four pivotal and representative VL tasks for our dataset:\footnote{See Appendix~\ref{sec:appendix-data-annotation} for the detailed annotation process. }

\textbf{Image-Text Retrieval}: This task includes text retrieval, where given an image, the task is to retrieve the corresponding text, and image retrieval, where given a text, the task is to retrieve the corresponding image. It evaluates VLMs' ability to align vision and language representations.

\textbf{Visual Question Answering}: Given an image and a natural language question, the model must generate a correct answer. It assesses VLMs' detailed visual understanding and reasoning skills.

\textbf{Visual Grounding}: Given an image and a referring expression, the model must locate the specified object. This task measures VLMs' ability to understand and identify objects in images.

\textbf{Visual Dialog}: Given an image, a dialogue history, and a question about the image, the model must answer accurately. This task evaluates VLMs' overall intelligence, including visual understanding, memory, and language generation.

\subsection{Image Collection}

After obtaining the list of object categories, our next goal was to collect appropriate images for each of them. 
To meet the requirements of different types of tasks in our dataset, we collect two subsets of images for each category. 
Subset A consists of images \textit{containing at least 2 objects of the same category} and is used for the VQA and VG tasks.\footnote{This constraint ensures VG is challenging enough.}
Subset B consists of images \textit{containing 3-5 objects of different object categories} and is used for the VD task.\footnote{This constraint improves the richness of dialogues in VD.}
The image captioning task is annotated on both subsets. 
All the collected images must be (1) real photos with no watermark; 
(2) non-iconic images with more than 2 objects; 
(3) commonly seen or representative in Chinese culture. 
The images were collected from the Chinese Internet and inspected by four co-authors who are well aware of the image collection guidelines. 

\subsection{Quality Control}

To ensure annotation quality, we use a two-step process for selecting and training annotators. 
First, candidates receive annotation guidelines and annotate five randomly sampled images to assess their general capability. 
Qualified candidates are then grouped by task based on their performance. 
Second, each group annotates 50 randomly sampled images, guided one-on-one by senior annotators until they fully understand the guidelines and achieve 100\% accuracy on these 50 images.


Annotators who completed the training began annotating tasks batched into packages. 
They could not proceed to the next package until finishing the current one. 
Each package was \textit{self-checked, reviewed by a senior inspector, and eventually inspected by four co-authors familiar with the guidelines}. 
The final inspection sampled 10\%-25\% of each package, requiring over 97\% accuracy to pass. 
Otherwise, the package was returned for correction. 
The IC, VQA, VG, and VD tasks involved 41, 108, 44, and 26 annotators and 10, 12, 8, and 13 senior inspectors, respectively. 
The project took six months and cost approximately RMB 550,000.

\section{Data Characteristics}

In this section, we analyse the annotated data to show their characteristics. 

\subsection{Images and Objects}

\begin{figure}[htbp]
  \centering
  \includegraphics[width=0.48\textwidth]{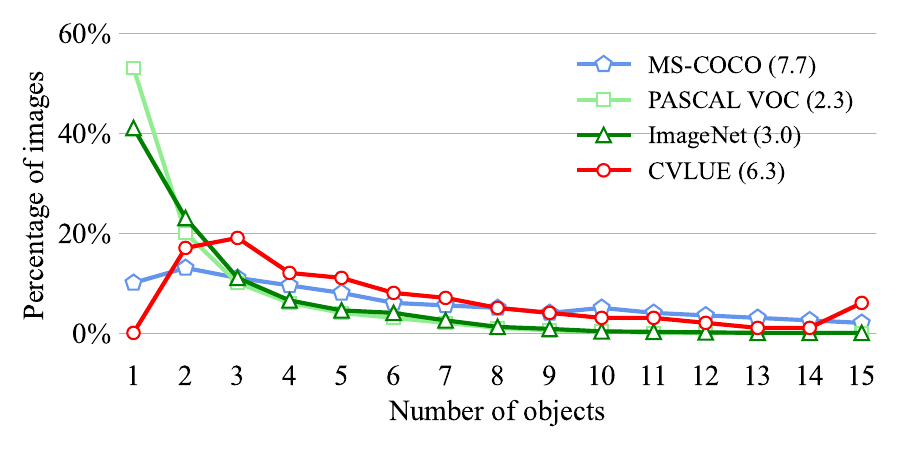}
  \caption{Number of annotated objects per image for CVLUE, MS-COCO, ImageNet Detection and PASCAL VOC (average numbers are shown in parentheses).}\label{fig:inst-per-img}
\end{figure}

We first count the object-related statistics to show the properties of the source images in CVLUE. 
The number of objects per category for all 92 categories is shown in Appendix~\ref{sec:appendix-cat-and-statistics}.  
We compare CVLUE with several popular datasets, including MS-COCO~\cite{lin-etal-2014-mscoco}, ImageNet\footnote{We use the object detection validation set since the training data only has a single object labelled.}~\cite{deng-etal-2009-imagenet} and PASCAL VOC~\cite{everingham-etal-2010-pascal-voc}. 
These datasets have different purposes: MS-COCO for detecting and segmenting objects in context, ImageNet for capturing object categories, and PASCAL VOC for detecting objects in natural images. 
CVLUE, however, is specifically designed to evaluate VLMs comprehensively in Chinese VLU.
Our dataset averages 6.3 annotated objects per image, compared to less than 3 for ImageNet and PASCAL VOC. Notably, no CVLUE images contain only one object due to subset A's requirement of at least two objects of the same category per image.
The numbers of annotated objects per image are shown in Figure~\ref{fig:inst-per-img}. 
Our dataset averages 6.3 annotated objects per image, compared to less than 3 for ImageNet and PASCAL VOC. 
Notably, no CVLUE images contain only one object due to subset A's requirement of at least two objects of the same category per image.

\subsection{Image Text Retrieval}

\begin{figure}[htbp]
  \centering
  \includegraphics[width=0.46\textwidth]{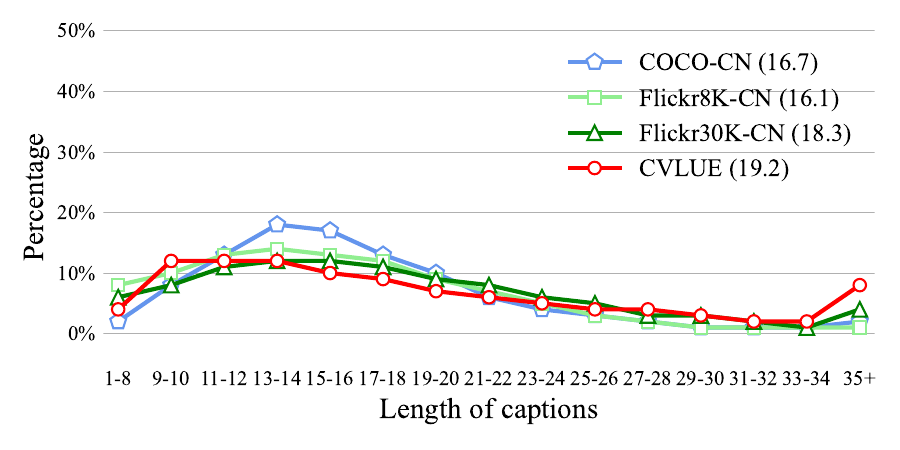}
  \caption{The caption length distribution of CVLUE, COCO-CN, Flickr8K-CN and Flickr30K-CN (average caption lengths are shown in parentheses).}
  \label{fig:cap-len-dist}
\end{figure}

\begin{table*}[ht]
	\centering
	\small
	\begin{tabular}{llccccc}
		\hline
		\multirow{3}{*}{Tasks} & \multirow{3}{*}{Dataset} & \multicolumn{2}{c}{Fine-tuning} & \multicolumn{3}{c}{Zero-shot}  \\
        \cmidrule(r){3-4} \cmidrule(r){5-7}
        &   & CCLM & X$^2$VLM & QwenVL & QwenVL-Chat &  mPLUG-Owl2  \\
        &   &  522M & 422M & 7B &  7B & 7B \\
		\hline
		\multirow{2}{*}{TR} & COCO (5K)   & 77.7 & 80.1 & - & - & - \\
         & CVLUE   & 49.9 & 54.8 & - & - & - \\
        \hline
        \multirow{2}{*}{IR} & COCO (5K)   & 60.5 & 63.8 & - & - & - \\
         & CVLUE   & 32.0 & 36.6 & - & - & - \\
        \hline
        \multirow{2}{*}{VQA} & VQA-v2 (test-std)  & 63.7 & 75.5 & 78.0 & 67.9 & 79.2 \\
        & CVLUE  & 58.5 & 53.0 & 29.9 & 39.8 & 20.4 \\
        \hline
        \multirow{2}{*}{VG} & RefCOCOg  & 70.4 & 79.9 & 78.0 & 80.1 & - \\
         & CVLUE  & 39.1 & 48.8 & 36.8 & 40.4 & - \\
        \hline
        \multirow{2}{*}{VD} & Visdial 1.0  & 42.4 & 41.5 & 36.0 & 37.5 & 37.2 \\
         & CVLUE   & 32.2 & 27.6 & 24.8 & 26.5 & 25.8 \\
		\hline
	\end{tabular}
	\caption{Results of baseline VLMs. 
We report R@1 for the TR, IR and VD tasks, accuracy for the VQA task and IoU for the VG task. 
For each compared model, we also report the number of parameters.}
	\label{tbl:main-result}
\end{table*}

For the ITR task, we compare CVLUE with several popular Chinese datasets constructed via text translation (Flickr30K) or re-annotation (Flickr8K and COCO-CN). 
These datasets are all built on top of Western culture-biased images from existing English VL datasets. 
The caption length distribution is shown in Figure~\ref{fig:cap-len-dist}. 
Our dataset's average caption length is 19.2, which is higher than that of the other three datasets. 
It is worth noting that the caption lengths in CVLUE are distributed more evenly than the other three datasets. 
This indicates that our dataset comprises both simple captions and complicated ones. 

\subsection{Visual Grounding}

To the best of our knowledge, there has not been any other Chinese VG dataset. 
To illustrate the property of the proposed dataset, here we provide a rough comparison between the VG dataset in CVLUE and a popular English VG dataset RefCOCOg \cite{mao-etal-2016-refcocog}. 
Overall, the average number of referring expressions per image is 3.38 for our VG dataset and 3.91 for RefCOCOg. 
This is because multiple expressions for a single object are allowed in RefCOCOg but disallowed in our dataset. 
The average number of objects described per image in our dataset and in RefCOCOg is 3.38 and 1.93, respectively, meaning that more objects are described in our dataset. 
Besides, the average expression lengths are 11.9 characters for our dataset and 8.3 words for RefCOCOg. 

\section{Experiments}

\subsection{Experimental Setups and Baselines}

We use CVLUE and some of its counterparts in English to evaluate the performance of several popular multilingual VLMs in VLU. 
The English VL datasets include COCO (5K) \cite{lin-etal-2014-mscoco}, VQA-v2 \cite{goyal-etal-2017-vqa2}, RefCOCOg \cite{mao-etal-2016-refcocog} and Visdial 1.0 \cite{das-etal-2017-visual}.\footnote{We use the default splits for these datasets.}

We use two experimental settings, namely the fine-tuning one and the zero-shot one. 
Models under the fine-tuning setting include:


\textbf{CCLM} \cite{zeng-etal-2023-cclm}, a multilingual VLM where the cross-lingual and cross-modal objectives are jointly learned. 

\textbf{X$^2$VLM} \cite{zeng-etal-2022-x2vlm}, a multilingual VLM where the multi-grained vision language alignments are learned in a unified framework. 

Models under the zero-shot setting include:

\textbf{Qwen-VL} \cite{bai-etal-2023-qwenvl}, a large-scale VLM pre-trained on 7 VL tasks simultaneously, can handle the grounding task. 

\textbf{Qwen-VL-Chat}, the Qwen-VL model fine-tuned through instruction tuning with the instruction following and dialogue capabilities enhanced. 

\textbf{mPLUG-Owl2} \cite{ye-etal-2023-owl2}, a large-scale VLM that incorporates shared functional modules to facilitate modality collaboration. 

We couldn't afford to tune hyper-parameters for each baseline model, so we used default ones for them all. 
Please refer to Appendix~\ref{sec:appendix-prompt} and \ref{sec:appendix-fine-tuning} for prompts used in the zero-shot setting and detailed fine-tuning setups. 
For the VD task, we collect 100 candidate answers (including correct, plausible, popular and random ones) for each question following the procedure proposed by \citet{das-etal-2017-visual}. 

\begin{figure*}[htbp]
  \centering
  \includegraphics[width=0.95\textwidth]{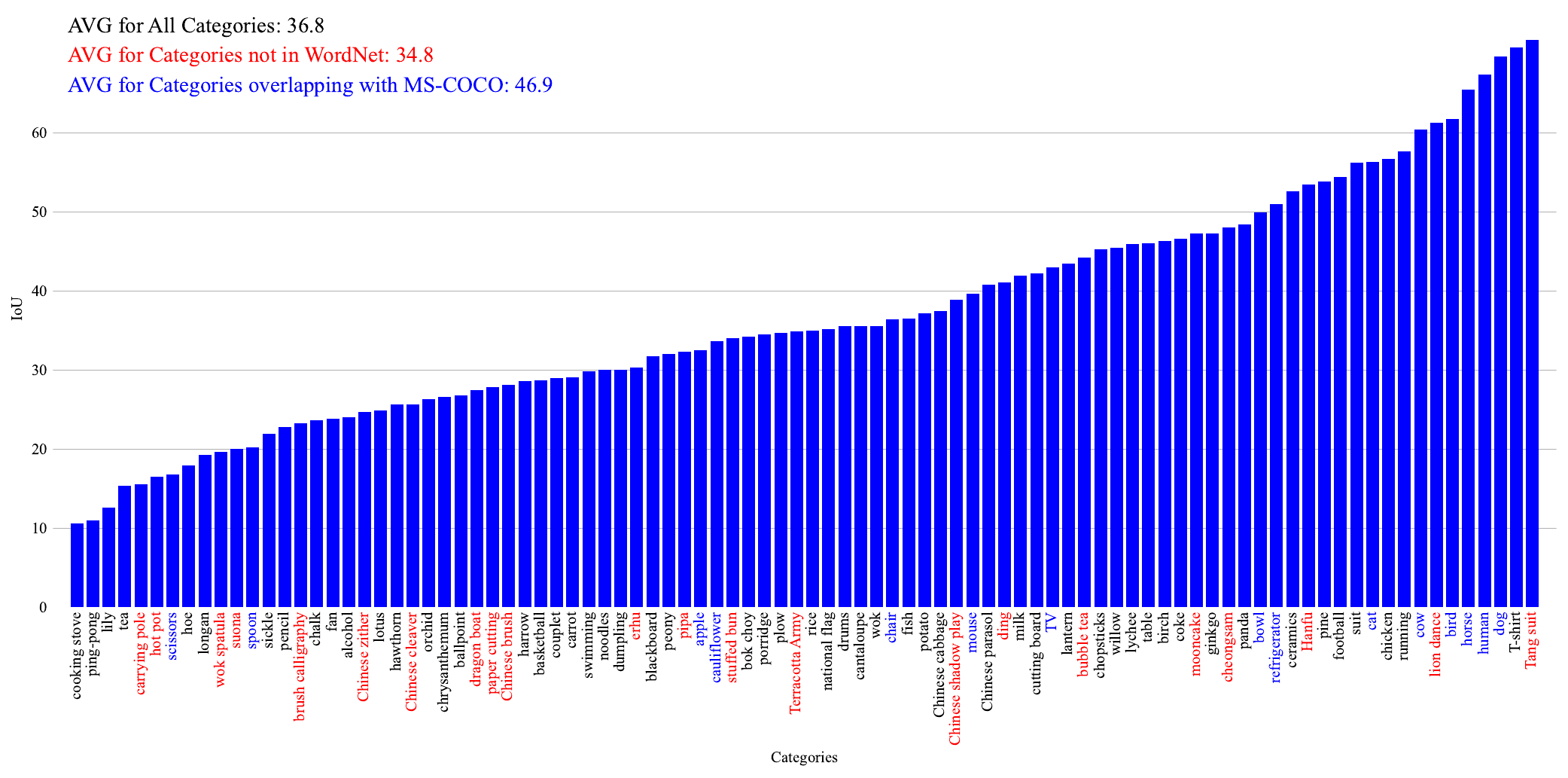}
  \caption{Results of QwenVL model on the CVLUE VG task, displayed by image category.}
  \label{fig:vg_qwen_by_cat}
\end{figure*}

\subsection{Results}

The results of the baseline models on CVLUE are presented in Table~\ref{tbl:main-result}.\footnote{See Appendix~\ref{sec:appendix-full-result} for full results containing R@5 and R@10 for the TR, IR and VD tasks.}
All models under the zero-shot setting do not support the ITR task. 
Additionally, mPLUG-Owl2 does not support the VG task either. 
Hence, these results are not reported.

The three large-scale VLMs under the zero-shot setting yield strong performance on the English datasets they are evaluated on, and some of their results are even higher than those of the two models under the fine-tuning setting. 
This could be attributed to their larger model capacity and the fact that they have been pre-trained on various VL tasks. 
On the other hand, all five models' performance on CVLUE is much lower than that on the English VL dataset. 
This aligns with the results observed on CCBench discussed in section~\ref{sec:related_work}.
Such a substantial performance gap between English and Chinese VL datasets indicates that the VLU capability of existing multilingual VLMs (under both zero-shot and fine-tuning settings) in Chinese severely lags behind that in English.
Besides, we find that on CVLUE, zero-shot models, despite having more parameters, often perform worse than fine-tuned models. Conversely, on English VL tasks, zero-shot models sometimes outperform fine-tuned ones. We believe this is because zero-shot models inherently possess more Western cultural knowledge than Chinese cultural knowledge, and their larger parameter scale gives them an edge in English tasks.

\section{Analysis}

\subsection{Results by Category}
To comprehensively investigate existing VLMs' VLU capabilities regarding Chinese culture, the first question to address is \textit{whether existing VLMs truly exhibit a significant performance difference between categories that are closely related to Chinese culture and those that are less related}.

Our dataset provides category information for each image, allowing for a fine-grained analysis of results across different categories. 
This facilitates the precise identification of the specific image categories in which VLMs exhibit deficiencies in their VLU abilities.
As discussed in section~\ref{sec:selection-of-object-categories}, the 92 categories in CVLUE can be roughly divided into three groups: 1) categories culturally closest to Chinese (i.e., those not in WordNet), 2) categories with the weakest association with Chinese culture (i.e., those overlapping with MS-COCO) and 3) categories with moderate association (i.e., the remaining ones). 
To answer the question, we analyze the models' results across different categories. 

Figure~\ref{fig:vg_qwen_by_cat} shows the performance of the QwenVL model on the VG task, displayed by category.
The results for categories closely related to Chinese culture are generally lower, with an average score of 34.8, while the results for categories overlapping with MS-COCO are generally higher, with an average score of 46.9.\footnote{Similar pattern observed on other tasks in Appendix~\ref{sec:appendix-result-by-cat}.}
This performance gap highlights a clear deficiency in existing VLMs' VLU capabilities regarding Chinese culture.

\subsection{Results on Translated English Test Sets}

Given that a majority of existing VL data used for pre-training focus on English with predominantly Western-centric images, the next question is \textit{whether the knowledge required to address tasks closely related to Chinese culture is present in the English part of existing VLMs}.

\begin{figure}[htbp]
  \centering
  \includegraphics[width=0.48\textwidth]{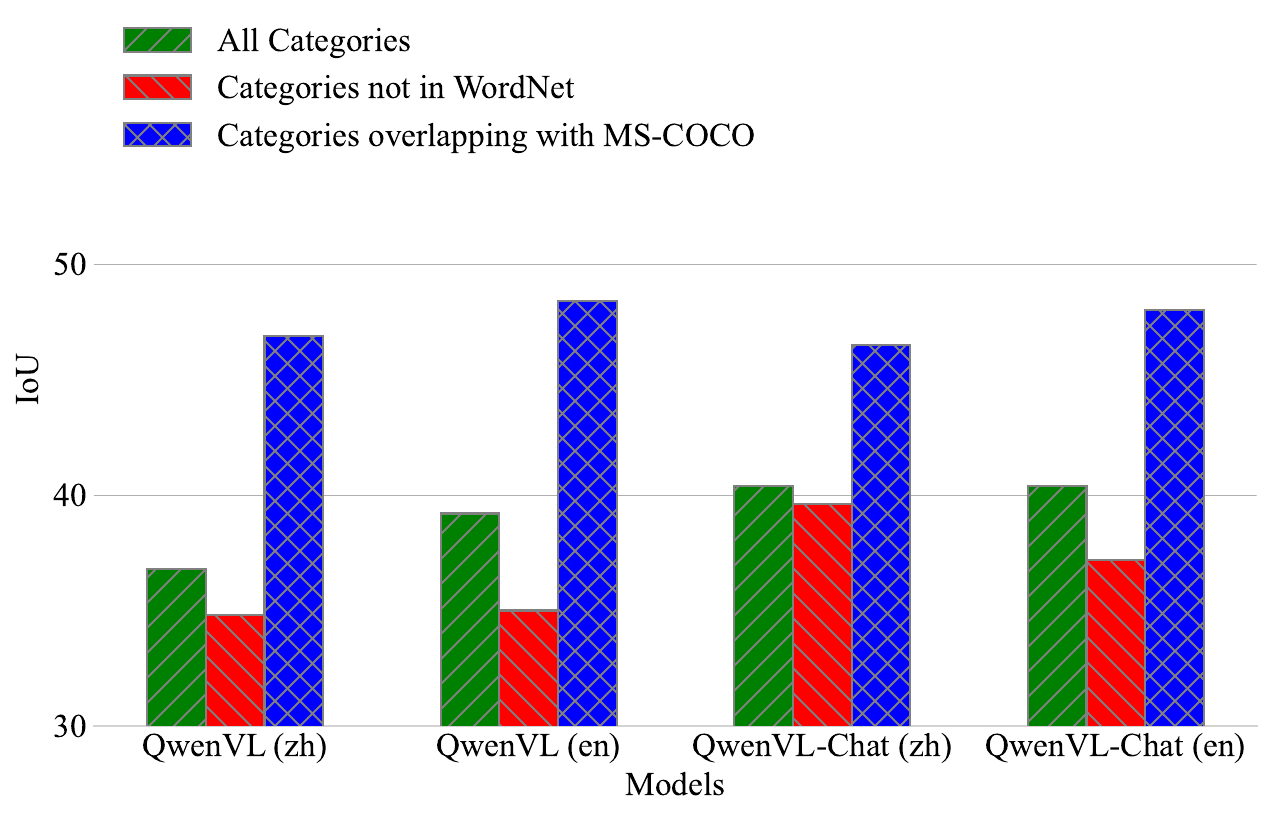}
  \caption{Category group results of QwenVL and QwenVL-Chat on the original Chinese (zh) and translated English (en) CVLUE VG test set.}
  \label{fig:vg_en_vs_zh_by_group}
\end{figure}

To address this question, we use GPT-4 to translate the VG test set into English, then compare QwenVL and QwenVL-Chat predictions with their results on the original Chinese test set. 
According to Figure~\ref{fig:vg_en_vs_zh_by_group}, for the same model, when the test set is translated from Chinese to English, performance on categories closely related to Chinese culture (not in WordNet) often remains unchanged or declines, while performance on categories less related to Chinese culture (overlapping with MS-COCO) significantly improves.\footnote{Similar pattern observed on VQA in Appendix~\ref{sec:appendix-result-on-english}.}
This indicates that in these VLMs, the English part typically contains more knowledge of categories less related to Chinese culture but, like the Chinese part, lacks knowledge of categories closely related to Chinese culture.

\subsection{Zero-Shot vs. Fine-Tuninig}

Due to the lack of knowledge required to address tasks closely related to Chinese culture in both the Chinese and English parts of existing VLMs, the final question becomes \textit{how to effectively enhance the knowledge of Chinese culture in these VLMs}.

\begin{figure}[htbp]
  \centering
  \includegraphics[width=0.48\textwidth]{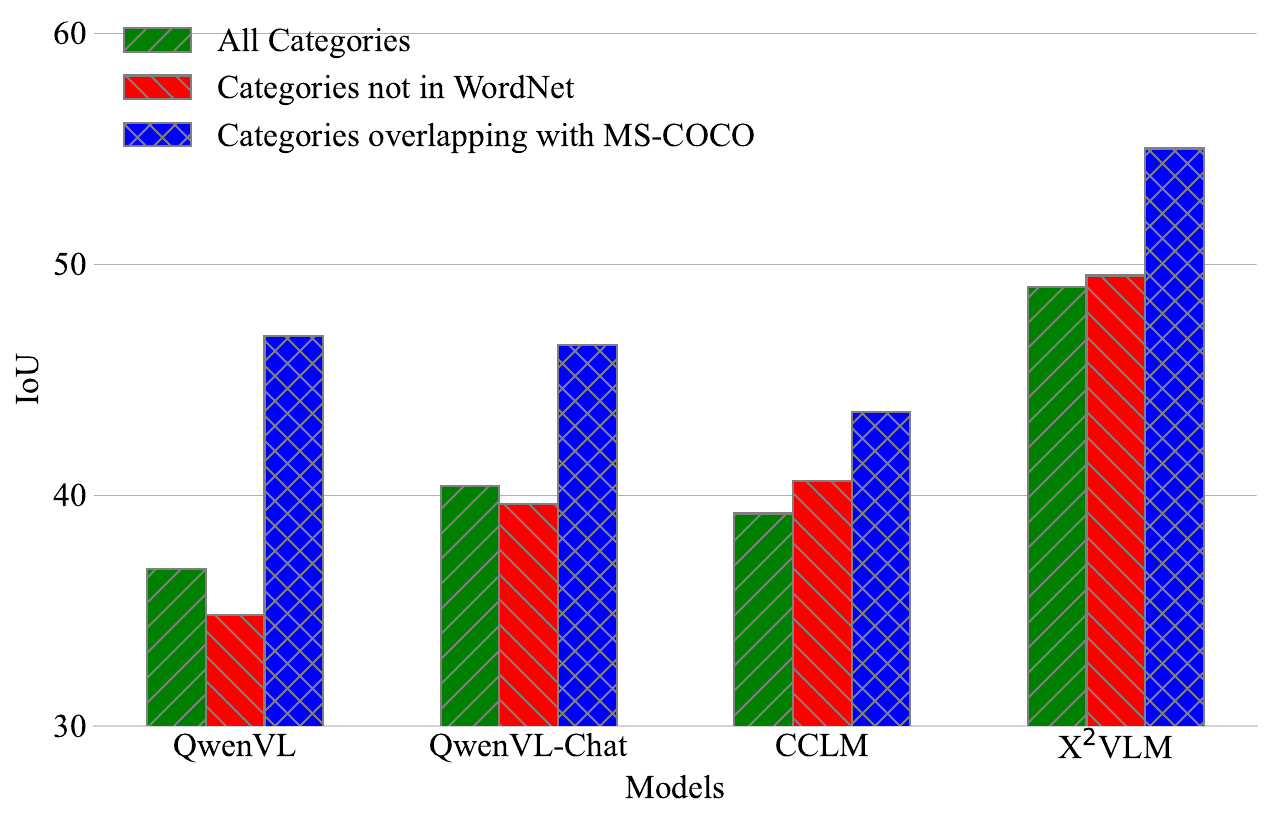}
  \caption{Category group results on CVLUE VG task.}
  \label{fig:vg_avg_by_group}
\end{figure}

In this section, we compare the performance of models under the zero-shot and the fine-tuning settings. 
According to the results on the CVLUE VG task in Figure~\ref{fig:vg_avg_by_group}, Chinese culture-related categories perform significantly lower than average on zero-shot models but higher than average on fine-tuned models.\footnote{Similar pattern observed on VQA in Appendix~\ref{sec:appendix-zs-vs-ft}.}
This indicates that fine-tuning with CVLUE's Chinese cultural VL data benefits categories strongly related to Chinese culture more. 
Overall, fine-tuning on Chinese cultural VL data is an effective way to enhance the VLM's VLU capabilities regarding Chinese culture.

\section{Conclusion}
\label{sec:conclusion}
In this paper, we present CVLUE, a vision-language understanding benchmark dataset specifically designed for the comprehensive evaluation of VLMs in Chinese VLU. 
Images used in the dataset were newly collected by Chinese native speakers with explicit constraints ensuring that they are representative of Chinese culture and thus avoid the cultural bias caused by exploiting images from existing English VL datasets. 
Four distinct and representative VL tasks are included in CVLUE for the multi-aspect evaluation of VLMs in Chinese culture. 
Using CVLUE and some English VL datasets, we reveal a noticeable gap between the performance of several strong multilingual VLMs on English and Chinese VLU. 
Our in-depth category-level analysis reveals a lack of Chinese culture-related knowledge in existing VLMs and shows that fine-tuning on Chinese culture-related VL datasets can effectively enhance VLMs' VLU capabilities regarding Chinese culture.
We believe that CVLUE is a solid step towards a fair and convenient platform for the comparison of VLMs in Chinese culture and can eventually facilitate the development of Chinese vision-language pre-training.

\section{Ethical Considerations}
\label{sec:ethical}

Images used in our benchmark are collected from the Chinese Internet. 
Sensitive information in the images (e.g., human faces) has been obscured to prevent potential misuse of the dataset. 
We used the Baidu data crowdsourcing platform for image collection and annotation. 
All the annotators have given informed consent and have been fairly compensated during the image collection and annotation process. 
The proposed dataset will be made publicly available for research purposes (under the CC BY-NC-ND 4.0 license). 

\section{Limitations}
\label{sec:limitation}

Due to limited computational resources, we were unable to test all VLMs or fine-tune larger VLMs on the proposed dataset. 
Therefore, we selected some popular and representative models and conducted experiments under both fine-tuning and zero-shot settings. 
Additionally, we couldn't afford to tune hyperparameters for each model, so we used the same default settings for all. 
Consequently, the reported results may not reflect the models' full potential. 
However, we believe that the current experimental setup is sufficient to highlight the significant performance gap between English and Chinese VL datasets for these strong and popular VLMs. 
Furthermore, the in-depth category-level analysis demonstrates that existing VLMs lack knowledge related to Chinese culture, validating the usefulness of CVLUE for comprehensive and fine-grained evaluation of VLMs in Chinese VLU.

Additionally, some may argue that the four tasks included in CVLUE are too few for a comprehensive evaluation of VLMs. 
However, due to budgetary constraints and to ensure both the quantity and quality of annotations, we could only select four important and representative VL tasks. 
Through in-depth, fine-grained analysis of the results on these tasks, we have found strong evidence that existing VLMs lack knowledge closely related to Chinese culture and proposed fine-tuning on Chinese cultural VL data as a solution to enhance VLMs' Chinese cultural VLU capabilities. 
Therefore, we believe that CVLUE is a solid step in the development of Chinese cultural VL benchmarks and hope it will inspire the creation of more extensive and comprehensive Chinese cultural VL datasets.

\bibliography{anthology,custom}
\bibliographystyle{acl_natbib}

\appendix
\newpage
\section{Appendix}
\label{sec:appendix}
\subsection{Categories and Statistics}
\label{sec:appendix-cat-and-statistics}

\begin{table}[ht]
	\centering
	\small
	\begin{tabular}{L{7cm}}
		\hline
		\bf Categories in MS-COCO \\
		\hline
            \blueit{person}, bicycle, car, motorcycle, airplane, bus, train, truck, boat, traffic light, 
            fire hydrant, street sign, stop sign, parking meter, bench, 
            \blueit{bird}, \blueit{cat}, \blueit{dog}, \blueit{horse}, sheep, \blueit{cow}, elephant, bear, zebra, giraffe, 
            hat, backpack, umbrella, shoe, eyeglasses, handbag, tie, suitcase, frisbee, skis, snowboard, sports ball, 
            kite, baseball bat, baseball glove, skateboard, surfboard, tennis racket, bottle, plate, wine glass, cup,
            fork, knife, \blueit{spoon}, \blueit{bowl}, banana, \blueit{apple}, sandwich, orange, \blueit{broccoli}, carrot,
            hot dog, pizza, donut, cake, \blueit{chair}, couch, potted plant, bed, mirror, dining table, window, desk, 
            toilet, door, \blueit{TV}, laptop, \blueit{mouse}, remote, keyboard, cell phone, microwave, oven, toaster, sink, 
            \blueit{refrigerator}, blender, book, clock, vase, \blueit{scissors}, teddy bear, hair drier, toothbrush, hairbrush \\
		\hline
	\end{tabular}
	\caption{Object categories in MS-COCO.}
	\label{tbl:coco-category}
\end{table}

The 91 object categories in MS-COCO, a popular English VL dataset and often used as the image source for other English and Chinese VL datasets, are listed in Table~\ref{tbl:coco-category}.

\begin{figure*}[htbp]
  \centering
  \includegraphics[width=0.95\textwidth]{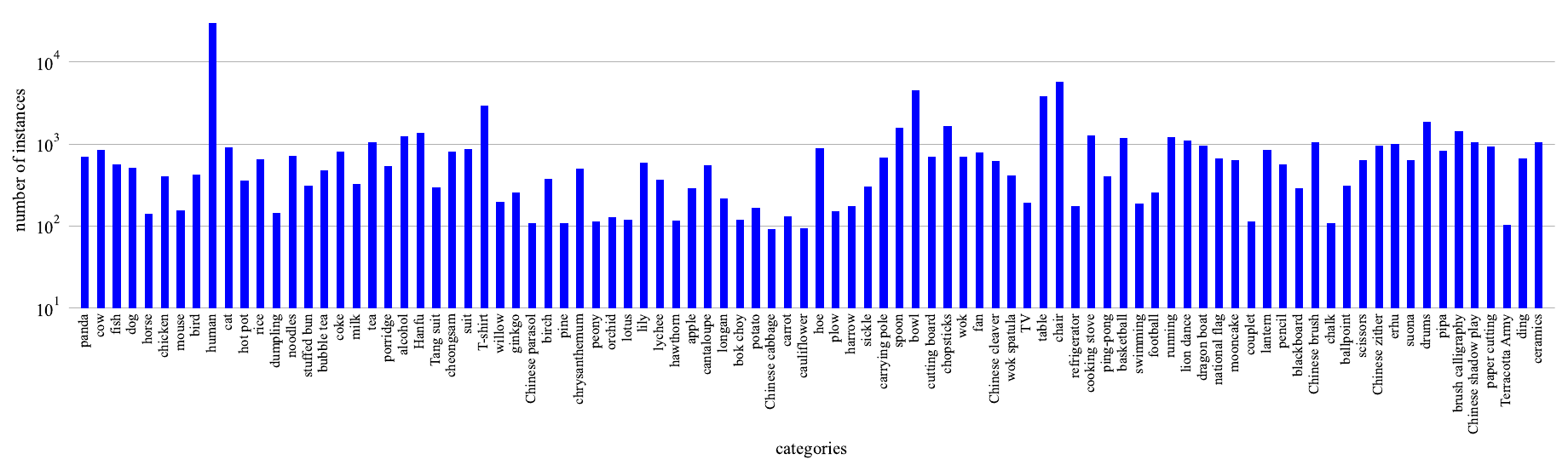}
  \caption{Number of annotated objects per category in CVLUE.}
  \label{fig:inst-per-cat}
\end{figure*}

The number of annotated objects per category for all 92 categories is shown in Figure~\ref{fig:inst-per-cat}. 

\subsection{Data Annotation}
\label{sec:appendix-data-annotation}

In this section, we introduce the detailed data annotation process for all the tasks in CVLUE. 

\subsubsection{Instance Segmentation}

The first stage is the task of segmenting object instances in images of subset A. 
All the objects belonging to the categories we selected above were manually labelled with bounding boxes.

\subsubsection{Image Captioning}

The image-text retrieval task includes two subtasks, namely text retrieval (TR), where given an image, the task is to retrieve the corresponding text and image retrieval (IR), where given a text, the task is to retrieve the image. 
This task aims to evaluate the capability of VLMs to align the semantic space of vision and language representations. 
The data is annotated via image captioning. 
Specifically, the annotators were asked to write five different sentences describing each image, which were required to:
\begin{itemize}
    \item Describe all the important parts of the image.
    \item Do not describe things that might have happened in the future or past.
    \item Do not describe what a person might say.
    \item Do not name people in the image.
    \item Contain at least eight characters. 
    \item Contain no more than 30\% overlapped characters between each other.
\end{itemize}

\subsubsection{Visual Question Answering}

Given an image and a natural language question, the VQA task requires the model to generate or select the corresponding answer in natural language. 
This task aims to evaluate VLMs' detailed visual understanding and complex reasoning ability.
Specifically, the annotators were asked to write three different questions for each image and give the correct answers in short phrases. 
The questions must: (1) require the image to correctly answer and not be answerable with only commonsense knowledge (e.g., `What is the book made of?'); and (2) not be too simple that only low-level computer vision knowledge is required to answer them (e.g., `What colour is the flower?'). 
The answers must be brief phrases rather than complete sentences. 
This constraint was added to ensure that the function of the VQA task is distinct from that of the VD task, in which the annotators were required to write complete sentences.

\subsubsection{Visual Grounding}

Given an image and a natural language referring expression, the VG task requires the model to locate the corresponding object. 
This task aims to evaluate the VLMs' ability to understand and distinguish objects in images. 
Specifically, each image was annotated by two annotators, namely A and B. 
A was asked to write an expression for each object labelled in the instance segmentation stage, distinguishing it from others of the same category.\footnote{For images containing more than four objects of the same category, we let the annotator select four objects to annotate.}
B was then given the expressions one by one and asked to select the corresponding object by clicking on the image. 
The annotation was regarded as correct only if B correctly selected all the objects.

An important factor that makes this task challenging enough is ensuring that at least two objects of the same category exist in all the images. 
Otherwise, this task would be degraded into simply distinguishing objects of different categories. 
\citet{kazemzadeh-etal-2014-referitgame} built their dataset on images from eixsting ImageCLEF dataset \cite{grubinger-etal-2006-iapr}. 
Therefore, they had no choice but to use images with and without multiple objects of the same category.
To deal with this issue, we restrict the number of objects of the same category from the beginning. 
Specifically, in the collection stage of subset A, we strictly require that only images containing at least two objects of the same category be included. 
Such categories will be considered as the \textit{main category} of the image. 
Then, during the VG annotation stage, the annotators were only asked to write expressions for the objects of the images' \textit{main category}. 
In this way, we guarantee that all the images used in this task contain two or more described objects of the same category, making the task more challenging.

\subsubsection{Visual Dialogue}

We employ the task of visual dialogue to evaluate the general intelligence of the VLM, ranging from global visual understanding to history memorization and natural language generation. 
The annotation of the VD task also requires the annotators to work in pairs. 
One of them was given a caption describing the image from subset B and was required to ask questions about the image to `imagine the scene better'. 
Another annotator was given both the image and the caption and was required to answer the questions based on the image. 
The conversation will be ended after ten pairs of questions and answers. 
It was emphasized to the annotators that the questions must be related to concrete objects in the image. 
Abstract questions concerning reason and meaning were not allowed. 

\subsection{Data Characteristics}

\subsubsection{Images and Objects}

\begin{figure}[htbp]
  \centering
  \includegraphics[width=0.48\textwidth]{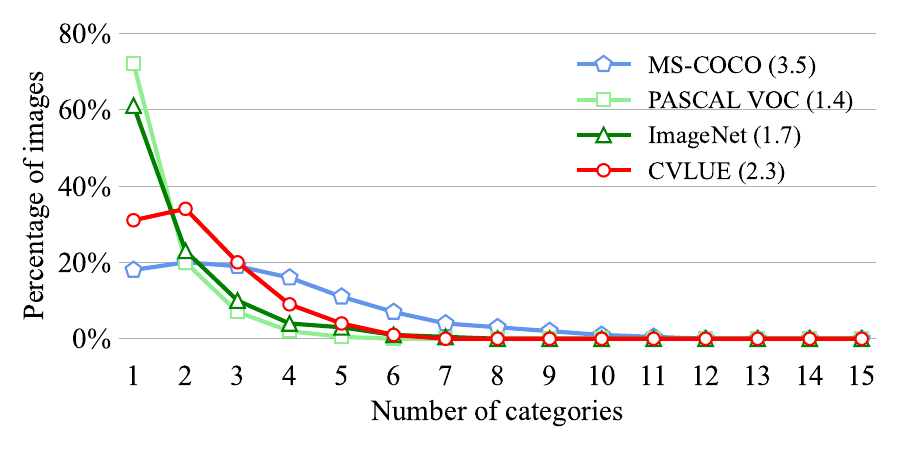}
  \caption{Number of annotated categories per image for CVLUE, MS-COCO, ImageNet Detection and PASCAL VOC (average number of categories are shown in parentheses).}\label{fig:cat-per-img}
\end{figure}

The numbers of annotated categories per image are shown in Figure~\ref{fig:cat-per-img}.

\subsubsection{Visual Question Answering and Visual Dialogue}
\label{sec:vqa-vd}

\begin{figure}[htbp]
  \centering
  \includegraphics[width=0.46\textwidth]{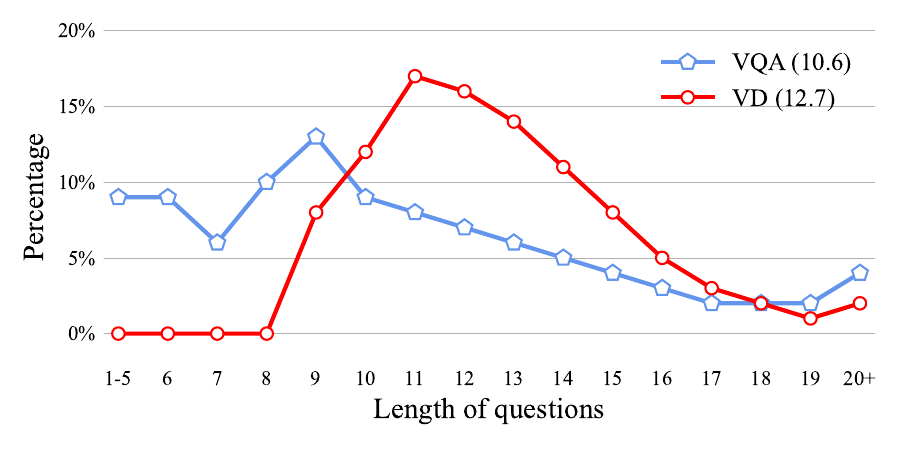}
  \caption{The question length distribution of VQA and VD in CVLUE (average lengths in parentheses).}
  \label{fig:quest-len-dist}
\end{figure}

\begin{figure}[htbp]
  \centering
  \includegraphics[width=0.46\textwidth]{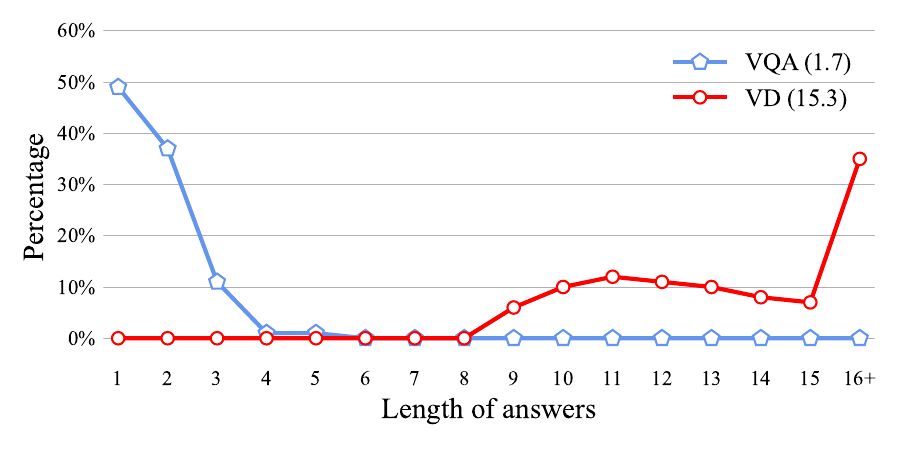}
  \caption{The answer length distribution of VQA and VD in CVLUE (average lengths in parentheses).}
  \label{fig:ans-len-dist}
\end{figure}

To illustrate the difference between VQA and VD tasks, we report their distribution of question and answer lengths in Figure~\ref{fig:quest-len-dist} and Figure~\ref{fig:ans-len-dist}, respectively. 
The question length distribution shows that VD has longer questions than VQA on average.
The difference becomes more evident in the answer length distribution, where answers in VQA are all short phrases, while VD has much longer answers. 

This difference reflects the distinct motivation of these two tasks. 
With VQA, we want the model to focus more on detailed visual understanding and complex reasoning. 
With VD, however, we want to evaluate VLMs' general intelligence, including global visual understanding, history memorization, and natural language generation. 
We also count the number of sentences containing pronouns (e.g., `he', `she', `it', etc.) and find that 43\% questions, 32\% answers and almost all (93\%) dialogues in VD contain at least one pronoun. 
In contrast, only 1\% of sentences in VQA contain pronouns. 
This means that the VD task also requires the capability to overcome coreference ambiguities, which is not strictly required by VQA.

To the best of our knowledge, there has not been any similar Chinese VD dataset. 
So, we make a rough comparison between the VD dataset in CVLUE and its English counterpart, the Visdial 1.0 dataset \cite{das-etal-2017-visual}.
We focus on the answers and find that the two most frequent answers for Visdial 1.0 are `no' and `yes', constituting 21.3\% and 19.2\% of the total answers, respectively. 
\begin{CJK}{UTF8}{gbsn}
For our VD dataset, the two most frequent answers are `这是一个女人/男人' (This is a woman/man), constituting only 0.1\% and 0.07\% of the total answers, respectively. 
\end{CJK}
Overall, Visdial 1.0 has 1,232,870 answers of 337,527 different types, while our VD dataset contains 97,550 answers of 93,308. 
The average answer lengths are 2.9 words for Visdial 1.0 and 15.3 characters for our VD dataset. 
This comparison shows our VD dataset's superiority regarding the answers' richness and complexity.

\subsection{CVLUE Examples}
\label{sec:appendix-cvlue-example}

\subsubsection{Image-Text Retrieval}

\begin{figure}[htbp]
  \centering
    \includegraphics[width=0.8\columnwidth]{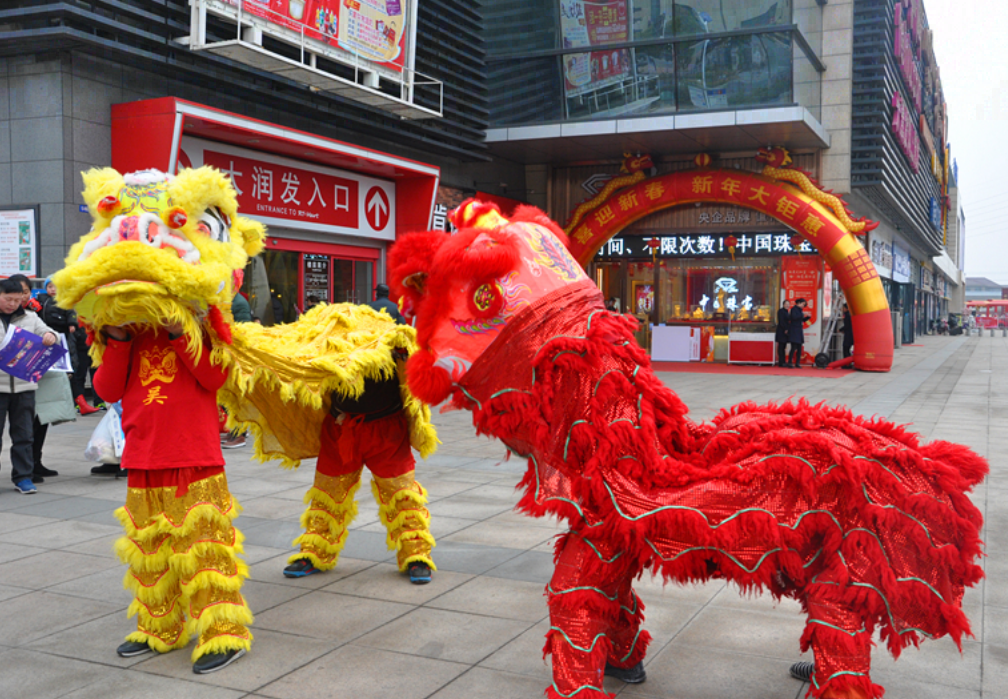}
    \caption{Image for ITR example 1.}
    \label{fig:itr-example-1}
\end{figure}

The 5 captions for ITR example 1 in Figure~\ref{fig:itr-example-1} are:
\begin{CJK}{UTF8}{gbsn}
\begin{itemize}
    \item 地面上有两只舞狮 (There are two dancing lions on the ground)
    \item 右边红色的舞狮横着站在地上，旁边还有一只黄色的舞狮 (On the right, a red dancing lion is standing horizontally on the ground, with a yellow dancing lion beside it)
    \item 有一只黄色的舞狮和一只红色的舞狮站在超市前边 (A yellow dancing lion and a red dancing lion are standing in front of a supermarket)
    \item 黄色的舞狮站起来了，旁边有另一只舞狮看着它，周围还有一些看客 (The yellow dancing lion is standing up, with another dancing lion watching it, surrounded by some spectators)
    \item 中国珠宝的店铺前有一个喜庆的拱门，前面有几只舞狮正在表演节目 (In front of a Chinese jewellery store, there is a festive archway, and several dancing lions are performing in front of it)
\end{itemize}
\end{CJK}

\begin{figure}[htbp]
  \centering
    \includegraphics[width=0.8\columnwidth]{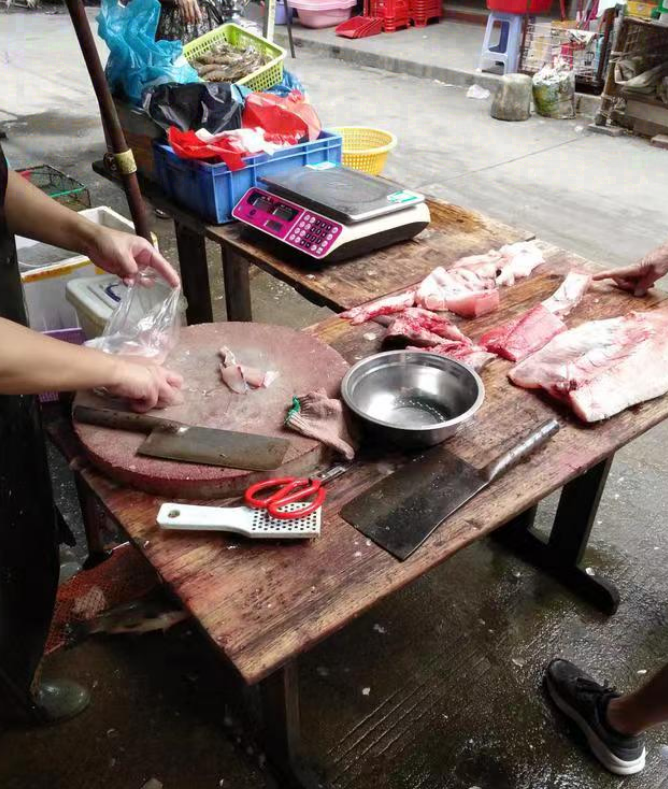}
    \caption{Image for ITR example 2.}
    \label{fig:itr-example-2}
\end{figure}

The 5 captions for ITR example 2 in Figure~\ref{fig:itr-example-2} are:
\begin{CJK}{UTF8}{gbsn}
\begin{itemize}
    \item 肉摊上的人在往塑料袋里面装肉，桌子上有菜刀 (The person at the meat stall is putting meat into a plastic bag, and there is a Chinese cleaver on the table)
    \item 电子秤旁边放着几块肉和一些菜刀，有个人在摊位前选肉 (Next to the electronic scale are some pieces of meat and Chinese cleavers, and a person is selecting meat at the stall)
    \item 一块圆形菜板上放着一些碎肉和一把菜刀，摊主手里提着塑料袋 (On a round cutting board, there are some pieces of meat and a Chinese cleaver, and the vendor is holding a plastic bag)
    \item 两块长方形桌子拼在一起，桌子上边有菜刀，下方有几个泡沫盒，有一条鱼在摊主的脚下 (Two rectangular tables are joined together, with Chinese cleavers on top and several foam boxes underneath, and there is a fish at the vendor's feet)
    \item 卖肉的摊位前有人经过，摊主的前面有菜刀和几块肉，选肉的人正用手指着其中一块 (Someone is passing by the meat stall; in front of the vendor are Chinese cleavers and pieces of meat, and the person selecting meat is pointing at one of the pieces)
\end{itemize}
\end{CJK}

\begin{figure}[htbp]
  \centering
    \includegraphics[width=0.8\columnwidth]{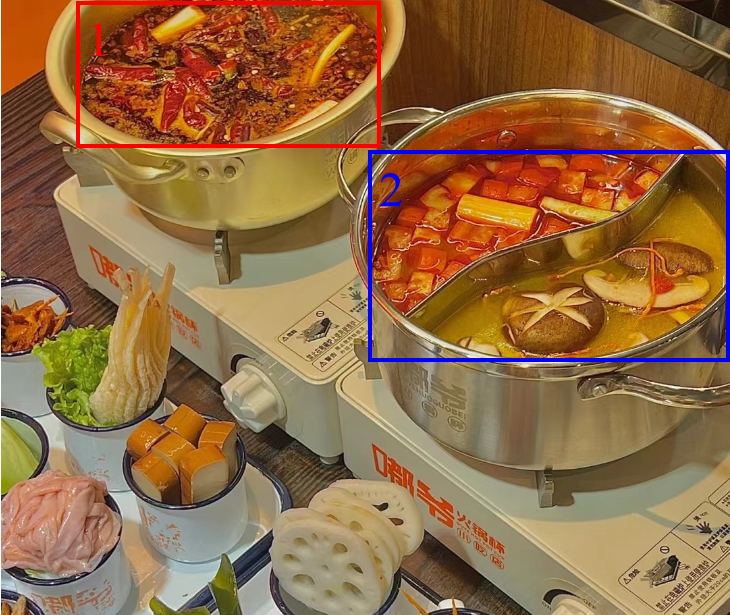}
    \caption{Image for ITR, VQA and VG example 3.}
    \label{fig:itr-vqa-vg-example-3}
\end{figure}

The 5 captions for ITR example 3 in Figure~\ref{fig:itr-vqa-vg-example-3} are:
\begin{CJK}{UTF8}{gbsn}
\begin{itemize}
    \item 桌上放着两口火锅和一些火锅食材 (There are two hot pots and some hot pot ingredients on the table)
    \item 两口火锅的下面放着用杯子装着的蔬菜和肉类 (Below the two hot pots, there are cups filled with vegetables and meat)
    \item 一些蔬菜和肉类的火锅食材被放在火锅的旁边，火锅下面放着加热灶 (Some vegetables and meat for the hot pot are placed beside the hot pot, and heating stoves are placed under the hot pots)
    \item 一口鸳鸯火锅和一口麻辣火锅放在了两个加热炉上，锅里面还放着一些食材 (A divided hot pot and a spicy hot pot are placed on two heating stoves, with some ingredients inside the pots)
    \item 装有食物的搪瓷杯子摆放在托盘上，食材旁边有两口火锅，它们分别放在了两台加热炉上 (Enamel cups filled with food are placed on a tray, and there are two hot pots beside the ingredients, each on a separate heating stove)
\end{itemize}
\end{CJK}

\subsubsection{Visual Question Answering}

\begin{figure}[htbp]
  \centering
    \includegraphics[width=0.8\columnwidth]{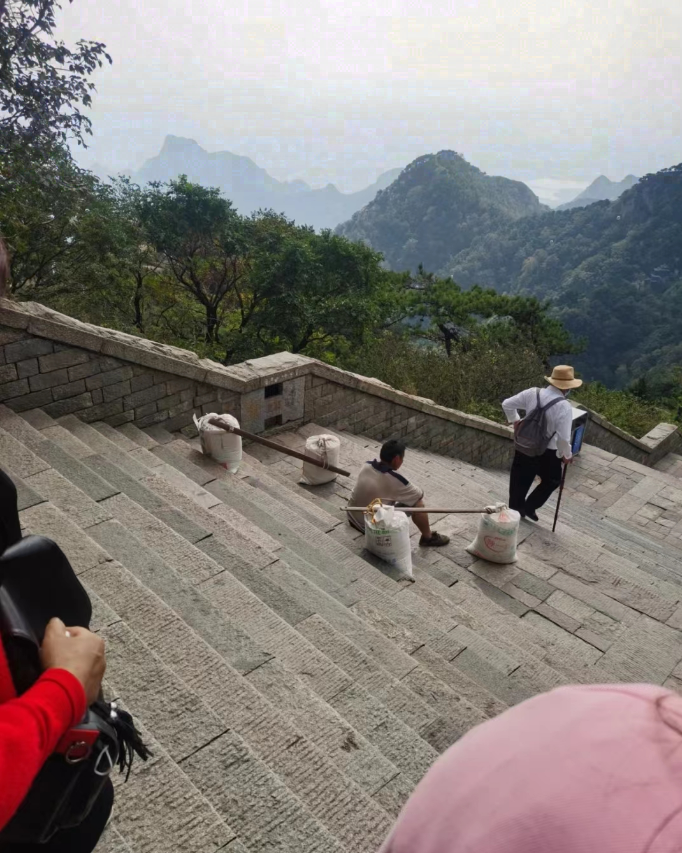}
    \caption{Image for VQA Example 1.}
    \label{fig:vqa-example-1}
\end{figure}

The 3 question-answer pairs for VQA example 1 in Figure~\ref{fig:vqa-example-1} are:

\begin{CJK}{UTF8}{gbsn}
\begin{itemize}
    \item Q: 戴帽子的人是在下台阶还是上台阶? (Is the person wearing a hat going down the steps or up the steps?) A: 下台阶 (Going down the steps)
    \item Q: 有几根担杖? (How many carrying poles are there?) A: 2
    \item Q: 两根担仗中间的人是坐着还是站着? (Is the person between the two carrying poles sitting or standing?) A: 坐着 (Sitting)
\end{itemize}
\end{CJK}

\begin{figure}[htbp]
  \centering
    \includegraphics[width=0.8\columnwidth]{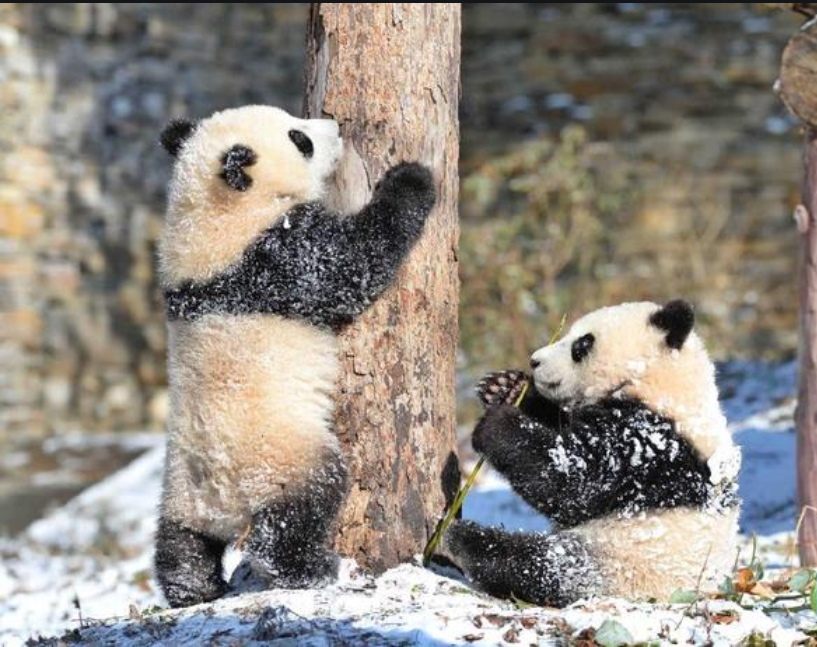}
    \caption{Image for VQA Example 2.}
    \label{fig:vqa-example-2}
\end{figure}

The 3 question-answer pairs for VQA example 2 in Figure~\ref{fig:vqa-example-2} are:

\begin{CJK}{UTF8}{gbsn}
\begin{itemize}
    \item Q: 当前是什么季节? (What season is it currently?) A: 冬季 (Winter)
    \item Q: 右侧大熊猫是什么姿势? (What is the posture of the panda on the right?) A: 坐着 (Sitting)
    \item Q: 坐着的大熊猫数量与站立的大熊猫数量相减等于几? (What is the difference between the number of sitting pandas and standing pandas?) A: 0
\end{itemize}
\end{CJK}

The 3 question-answer pairs for VQA example 3 in Figure~\ref{fig:itr-vqa-vg-example-3} are:

\begin{CJK}{UTF8}{gbsn}
\begin{itemize}
    \item Q: 圆形切片的是什么食材? (What ingredient is the round slice?) A: 藕 (Lotus root)
    \item Q: 火锅的口味相同吗? (Are the flavors of the hot pots the same?) A: 不相同 (No)
    \item Q: 鸳鸯锅里褐色食材是什么? (What is the brown ingredient in the divided hot pot?) A: 香菇 (Shiitake mushrooms)
\end{itemize}
\end{CJK}

\subsubsection{Visual Grounding}

\begin{figure}[htbp]
  \centering
    \includegraphics[width=0.8\columnwidth]{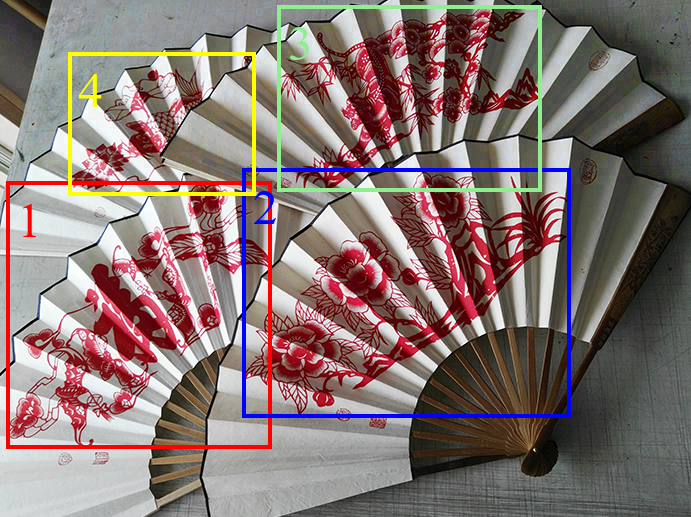}
    \caption{Image for VG Example 1.}
    \label{fig:vg-example-1}
\end{figure}

The 4 referring expressions for VG example 1 in Figure~\ref{fig:vg-example-1} are:

\begin{CJK}{UTF8}{gbsn}
\begin{itemize}
    \item 1.扇子上有汉字的剪纸 (The paper cutting on the fan in the shape of Chinese characters)
    \item 2.扇子上有三朵花形状的剪纸 (The paper cutting on the fan in the shape of three flowers)
    \item 3.有老虎形状的红色剪纸 (The red paper cutting in the shape of a tiger)
    \item 4.扇子上是荷叶和鱼形状的剪纸 (The paper cutting on the fan in the shape of lotus leaves and fish)
\end{itemize}
\end{CJK}

\begin{figure}[htbp]
  \centering
    \includegraphics[width=0.8\columnwidth]{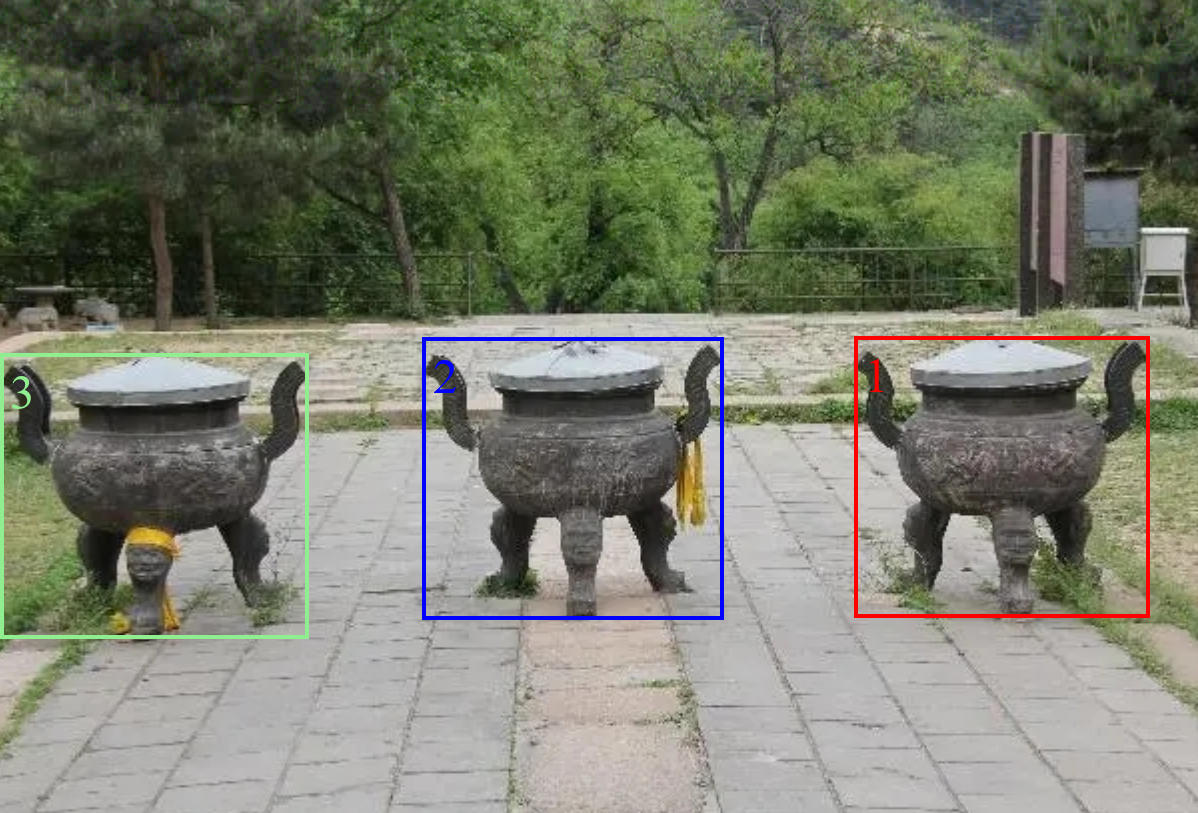}
    \caption{Image for VG Example 2.}
    \label{fig:vg-example-2}
\end{figure}

The 3 referring expressions for VG example 2 in Figure~\ref{fig:vg-example-2} are:

\begin{CJK}{UTF8}{gbsn}
\begin{itemize}
    \item 1: 鼎身没有系黄色绸带的一只鼎 (The ding without a yellow silk ribbon tied around its body)
    \item 2: 位于中间的一只鼎 (The ding in the middle)
    \item 3: 鼎脚上系了一根黄色绸带的一只鼎 (The ding with a yellow silk ribbon tied around its leg)   
\end{itemize}
\end{CJK}

The 2 referring expressions for VG example 3 in Figure~\ref{fig:itr-vqa-vg-example-3} are:

\begin{CJK}{UTF8}{gbsn}
\begin{itemize}
    \item 1: 汤汁上飘有许多干辣椒的那个火锅 (The hot pot with many dried chili peppers floating on the broth)
    \item 2: 装有香菇的那个鸳鸯火锅 (The divided hot pot with shiitake mushrooms)
\end{itemize}
\end{CJK}

\subsubsection{Visual Dialogue}

\begin{figure}[htbp]
  \centering
    \includegraphics[width=0.8\columnwidth]{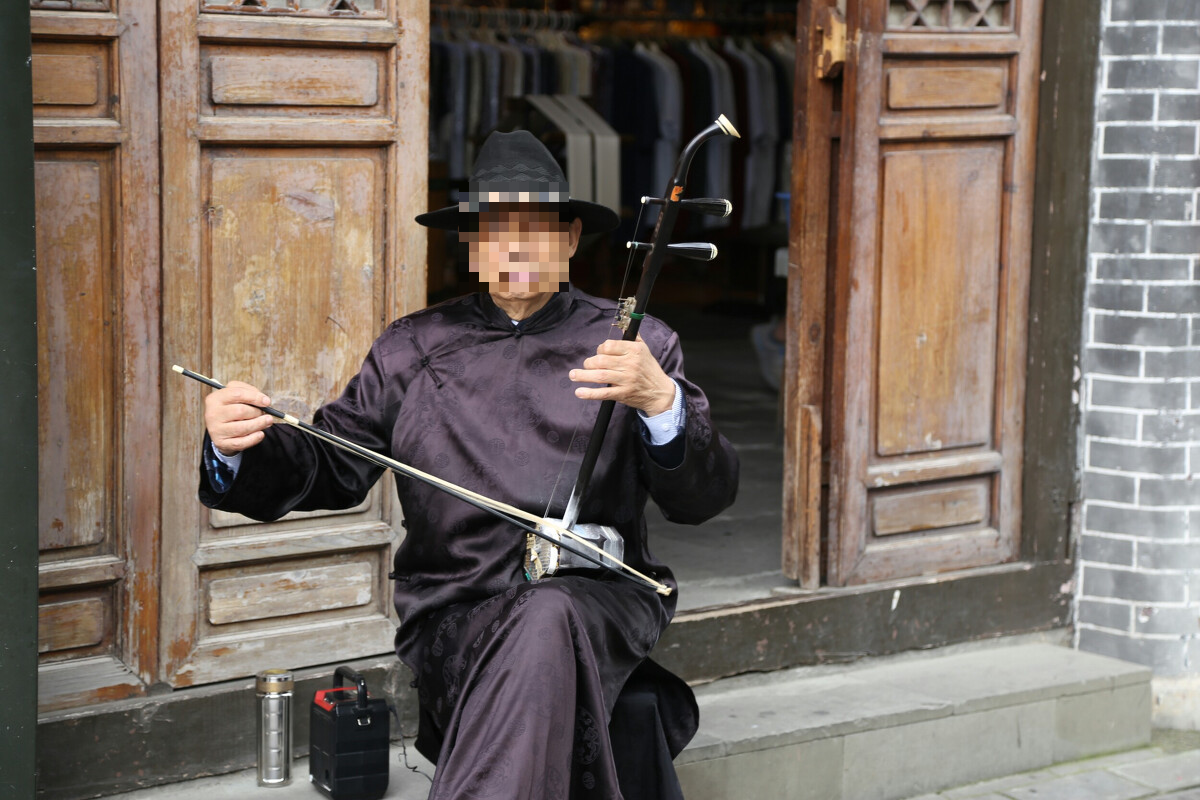}
    \caption{Image for VD Example 1.}
    \label{fig:vd-example-1}
\end{figure}

The caption and 10 rounds of dialogue for VD example 1 in Figure~\ref{fig:vd-example-1} are:

\begin{CJK}{UTF8}{gbsn}
\begin{itemize}
    \item Caption: 戴着黑色帽子的男人在拉二胡 (A man wearing a black hat is playing the erhu)
    \item Q: 这个男人穿着什么样的衣服? (What kind of clothes is this man wearing?) A: 这个男人穿着黑色的长袍 (This man is wearing a black robe)
    \item Q: 这个男人是在室内还是室外? (Is this man indoors or outdoors?) A: 这个男人是在室外 (This man is outdoors)
    \item Q: 这个男人周围都有什么? (What is around this man?) A: 这个男人旁边有音响和水杯, 背后有门和墙 (Next to this man, there are a speaker and a water cup, and behind him, there is a door and a wall)
    \item Q: 这个男人背后的墙是什么样式的? (What is the style of the wall behind this man?) A: 这个男人背后的墙是灰色的砖块砌成的 (The wall behind this man is made of grey bricks)
    \item Q: 这个男人背后的门是什么颜色的? (What colour is the door behind this man?) A: 这个男人背后的门是棕黄色的 (The door behind this man is brownish-yellow)
    \item Q: 男人背后的门是什么材质的? (What material is the door behind the man made of?) A: 男人背后的门是木头制作的 (The door behind the man is made of wood)
    \item Q: 男人背后的门是打开的吗? (Is the door behind the man open?) A: 男人背后的门是打开的 (The door behind the man is open)
    \item Q: 男人旁边的音响是什么颜色的? (What colour is the speaker next to the man?) A: 男人旁边的音响是黑色的 (The speaker next to the man is black)
    \item Q: 男人旁边的水杯是什么颜色的? (What colour is the water cup next to the man?) A: 男人旁边的水杯是银白色的 (The water cup next to the man is silver)
    \item Q: 这个水杯是盖着的吗? (Is the water cup covered?) A: 这个水杯是盖着的 (The water cup is covered)
\end{itemize}
\end{CJK}

\begin{figure}[htbp]
  \centering
    \includegraphics[width=0.8\columnwidth]{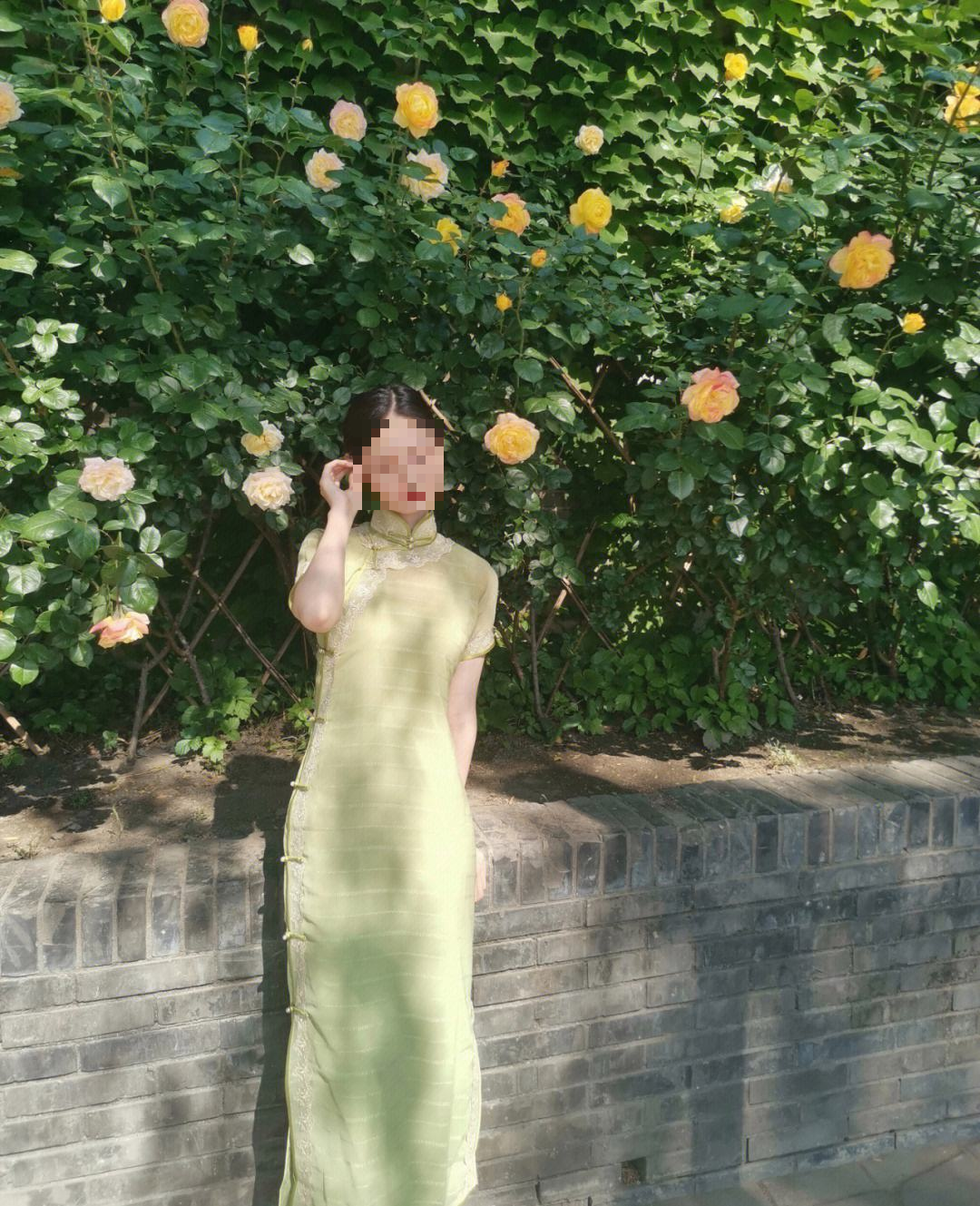}
    \caption{Image for VD Example 2.}
    \label{fig:vd-example-2}
\end{figure}

The caption and 10 rounds of dialogue for VD example 2 in Figure~\ref{fig:vd-example-2} are:

\begin{CJK}{UTF8}{gbsn}
\begin{itemize}
    \item Caption: 砖墙前倚靠着一位旗袍女子 (A woman in a cheongsam leaning against a brick wall)
    \item Q: 图中的砖墙是什么颜色的呢？ (What colour is the brick wall in the picture?) A: 图中的砖墙是浅灰色的砖墙。 (The brick wall in the picture is light grey.)
    \item Q: 图中的女子穿着一件什么颜色的旗袍呢？ (What colour is the cheongsam the woman is wearing?) A: 图中的女子穿了一件浅绿色的旗袍。 (The woman in the picture is wearing a light green cheongsam.)
    \item Q: 图中女子穿的旗袍是长袖还是短袖的呢？ (Is the cheongsam the woman is wearing long-sleeved or short-sleeved?) A: 图中的女子穿的旗袍是短袖的。 (The cheongsam the woman is wearing is short-sleeved.)
    \item Q: 图中女子的动作是什么样子的呢？ (What is the woman doing in the picture?) A: 女子倚靠着砖墙，右手抚摸着耳朵眼睛看着镜头。 (The woman is leaning against the brick wall, touching her ear with her right hand and looking at the camera.)
    \item Q: 图中女子是在室内还是室外呢？ (Is the woman indoors or outdoors in the picture?) A: 女子是在室外，她的身后还有很多花。 (The woman is outdoors, with many flowers behind her.)
    \item Q: 女子身后的花是什么颜色的呢？ (What colour are the flowers behind the woman?) A: 女子身后的花有浅粉色的和橘黄色的。 (The flowers behind the woman are light pink and orange.)
    \item Q: 图中女子看上去年龄有多大呢？ (How old does the woman in the picture look?) A: 图中的女子看上去很年轻，二十来岁。 (The woman in the picture looks very young, in her twenties.)
    \item Q: 图中除了女子，还有其他的人吗？ (Are there any other people in the picture besides the woman?) A: 图中只有女子一个人，没有其他的人。 (There is only the woman in the picture, no one else.)
    \item Q: 图中的天气情况怎么样呢？ (What is the weather like in the picture?) A: 图中阳光明媚，是一个晴天。 (The weather in the picture is sunny and bright.)
    \item Q: 图中的女子身上有什么装饰品吗？ (Is the woman wearing any accessories in the picture?) A: 图中的女子头上带了一个浅棕色的发饰。 (The woman in the picture is wearing a light brown hair accessory.)
\end{itemize}
\end{CJK}

\begin{figure}[htbp]
  \centering
    \includegraphics[width=0.8\columnwidth]{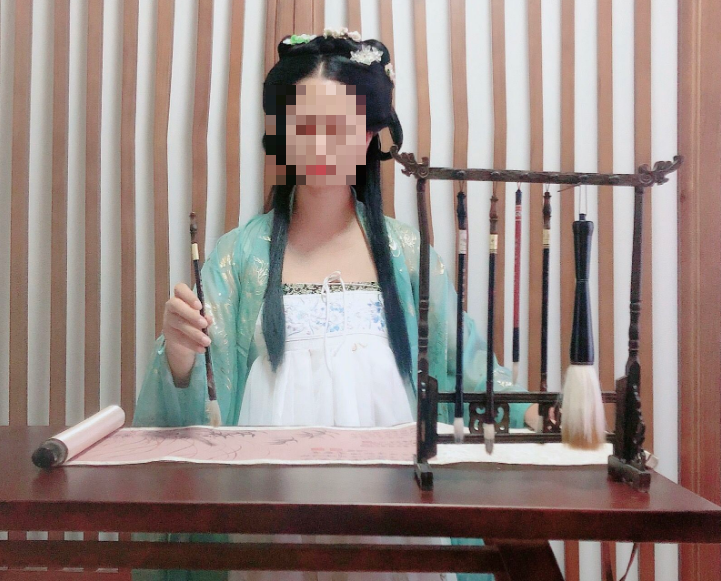}
    \caption{Image for VD Example 3.}
    \label{fig:vd-example-3}
\end{figure}

The caption and 10 rounds of dialogue for VD example 3 in Figure~\ref{fig:vd-example-3} are:

\begin{CJK}{UTF8}{gbsn}
\begin{itemize}
    \item Caption: 屋内坐着一个穿着汉服的女生 (A girl in Hanfu sitting indoors)
    \item Q: 她的旁边有什么东西？ (What is next to her?) A: 有一个桌子和毛笔架。 (There is a table and a Chinese brush rack.)
    \item Q: 毛笔架上面有毛笔吗？ (Are there brushes on the brush rack?) A: 是的，上面挂着五支毛笔。 (Yes, there are five Chinese brushes hanging on it.)
    \item Q: 她穿的汉服是什么颜色的？ (What colour is the Hanfu she is wearing?) A: 浅绿色的有金色的图案。 (It is light green with golden patterns.)
    \item Q: 她的手在什么位置？ (Where are her hands?) A: 右手握着毛笔，左手放在纸上面。 (Her right hand is holding a Chinese brush, and her left hand is on the paper.)
    \item Q: 画纸上面有镇纸吗？ (Is there a paperweight on the drawing paper?) A: 没有，桌子上只有画纸和毛笔架。 (No, there is only drawing paper and the Chinese brush rack on the table.)
    \item Q: 她的旁边没有其他人吗？ (Is there no one else next to her?) A: 是的，只有她一个人。 (Yes, she is alone.)
    \item Q: 毛笔架上面的毛笔都长得一样吗？ (Do all the brushes on the brush rack look the same?) A: 不是，每个毛笔的样式都不一样。 (No, each Chinese brush is different.)
    \item Q: 桌子是什么颜色的？ (What colour is the table?) A: 是深红色的与黑色混杂的桌面。 (It is a dark red and black mixed tabletop.)
    \item Q: 她的后面有什么东西？ (What is behind her?) A: 是一些木质的架子，原木的颜色。 (There are some wooden shelves, natural wood colour.)
    \item Q: 她的头上有什么装饰品？ (Is she wearing any accessories on her head?) A: 有绿色的和白色的小花。 (She has small green and white flowers in her hair.)
\end{itemize}
\end{CJK}

\subsection{Fine-tuning Experimental Setups}
\label{sec:appendix-fine-tuning}
In the fine-tuning setting, all tasks use the AdamW optimizer with a weight decay of 0.05 and the cosine learning rate scheduler. 
We use the default image resolution for each of the baseline models. 
Other hyper-parameters are listed in Table~\ref{tbl:hyper}. 
In the fine-tuning setting, during the inference stage of VQA, we constrain the decoder to only generate from candidates computed in the training and valid set. 
The models were fine-tuned on 8 V100s. 

\begin{table}[ht]
	\centering
	\small
	\begin{tabular}{l|cccc}
		\hline
		Task & init LR & batch size & resolution & \#epoch \\
		\hline
        ITR & $3e^{-5}$ & 128 & 384$\times$384 & 10 \\
        VQA & $3e^{-5}$ & 128 & 768$\times$768 & 5 \\
        VG &  $1e^{-5}$ & 128 & 384$\times$384 & 10 \\
        VD &  $3e^{-5}$ & 128 & 384$\times$384 & 5 \\
		\hline
	\end{tabular}
	\caption{Hyper-parameters used in the fine-tuning setting. init LR stands for initial learning rate. }
	\label{tbl:hyper}
\end{table}

\subsection{Experimental Results}
\label{sec:appendix-full-result}

The data splits of the English VL datasets we used are shown in Table~\ref{tbl:en-split}. 

\begin{table}[ht]
	\centering
	\small
	\begin{tabular}{l|ccc}
		\hline
		Task & $|$Train$|$ & $|$Valid$|$ & $|$Test$|$  \\
		\hline
        COCO (5K) & 82,783 & 5,000 & 5,000  \\
        VQA-v2 & 82,783 & 40,504 & 81,434  \\
        RefCOCOg  & 21,899 & 1,300 & 2,600  \\
        Visdial 1.0  & 123,287  & 2,064  & 8,000 (QA pairs) \\
		\hline
	\end{tabular}
	\caption{Data splits (in terms of image numbers if not explicitly specified) of the English VL datasets we used. }
	\label{tbl:en-split}
\end{table}

\begin{table*}[htbp]
	\centering
	\small
	\begin{tabular}{lllccccc}
		\hline
		\multirow{3}{*}{Tasks} & \multirow{3}{*}{Dataset} & \multirow{3}{*}{Metrics} & \multicolumn{2}{c}{Fine-tuning} & \multicolumn{3}{c}{Zero-shot}  \\
        \cmidrule(r){4-5} \cmidrule(r){6-8}
        &  &  & CCLM & X$^2$VLM & QwenVL & QwenVL-Chat &  mPLUG-Owl2  \\
        &  &  &  522M & 422M & 7B &  7B & 7B \\
		\hline
		\multirow{6}{*}{TR} & \multirow{3}{*}{COCO (5K)}  
          & R@1   & 77.7 & 80.1 & - & - & - \\
         & & R@5  & 94.2 & 95.3 & - & - & - \\
         & & R@10 & 97.1 & 97.6 & - & - & - \\
         \cline{2-8}
         & \multirow{3}{*}{CVLUE} & R@1  & 49.9 & 54.8 & - & - & - \\
         & & R@5  & 75.2 & 79.5 & - & - & - \\
         & & R@10 & 82.8 & 86.8 & - & - & - \\
        \hline
        \multirow{6}{*}{IR} & \multirow{3}{*}{COCO (5K)}  
          & R@1   & 60.5 & 63.8 & - & - & - \\
         & & R@5  & 84.3 & 86.1 & - & - & - \\
         & & R@10 & 90.7 & 91.8 & - & - & - \\
         \cline{2-8}
         & \multirow{3}{*}{CVLUE} & R@1  & 32.0 & 36.6 & - & - & - \\
         & & R@5  & 58.3 & 63.4 & - & - & - \\
         & & R@10 & 69.6 & 73.6 & - & - & - \\
        \hline
        \multirow{2}{*}{VQA} & VQA-v2 (test-std)  & Acc & 63.7 & 75.5 & 78.0 & 67.9 & 79.2 \\
        & CVLUE & Acc & 58.5 & 53.0 & 29.9 & 39.8 & 20.4 \\
        \hline
        \multirow{2}{*}{VG} & RefCOCOg & IoU & 70.4 & 79.9 & 78.0 & 80.1 & - \\
         & CVLUE & IoU & 39.1 & 48.8 & 36.8 & 40.4 & - \\
        \hline
        \multirow{6}{*}{VD} & \multirow{3}{*}{Visdial 1.0}  
          & R@1   & 42.4 & 41.5 & 36.0 & 37.5 & 37.2 \\
         & & R@5  & 64.4 & 59.7 & 50.0 & 51.8 & 52.4\\
         & & R@10 & 72.5 & 67.7 & 55.6 & 57.6 & 59.4 \\
         \cline{2-8}
         & \multirow{3}{*}{CVLUE} & R@1  & 32.2 & 27.6 & 24.8 & 26.5 & 25.8 \\
         & & R@5  & 46.6 & 41.0 & 34.9 & 35.9 & 38.3 \\
         & & R@10 & 53.3 & 47.8 & 39.9 & 40.2 & 45.3 \\
		\hline
	\end{tabular}
	\caption{Results of baseline VLMs. R@1, R@5 and R@10 denote the recall in the top 1, 5 and 10 predictions, respectively. Acc denotes the accuracy, and IoU stands for the average intersection over union. For each compared model, we also report the number of parameters.}
	\label{tbl:full-result}
\end{table*}

The full experimental results are shown in Table~\ref{tbl:full-result}. 

\subsection{Prompts for the Zero-Shot Setting}
\label{sec:appendix-prompt}

\subsubsection{Visual Question Answering}
\label{sec:appendix-vqa-prompt}
In the VQA task, we use the prompts \begin{CJK}{UTF8}{gbsn}`用尽量简洁的数字或中文短语回答以下问题：[question]'\end{CJK} for Chinese and `Answer the question with only an Arabic figure or a phrase: [question]' for English, where [question] denotes the question in VQA. 

\subsubsection{Visual Grounding}

In the VG task, we use the prompts \begin{CJK}{UTF8}{gbsn}`框出图中[expression]的位置'\end{CJK} for Chinese and `<ref>[expression]</ref><box>' for English, where [expression] denotes the referring expression in VG, <ref>, </ref> and <box> are special tokens in the Qwen-VL model. 

\subsubsection{Visual Dialogue}

In the VD task, we use the prompts \begin{CJK}{UTF8}{gbsn}`描述: [caption] 对话历史: [history] 根据图片描述和对话历史用一句话回答以下问题. 问题: [question] 答案:'\end{CJK} for Chinese and `Context: [caption] History: [history] Answer the question with one sentence based on the context and dialogue history. Question:  [question] Answer:' for English. 
[caption] denotes the caption describing the image in VD, [history] denotes the dialogue history, which is also in the format of question-answer pairs, and [question] denotes the current question to be answered in this round of dialogue.

Since the VD task is to rank the 100 answer candidates given the dialogue history and current question, we could not directly apply the generative VLMs in such a situation. 
Therefore, we concatenate each answer candidate with the dialogue history and the current question and use the VLM to calculate their probabilities, eventually ranking all candidate answers based on these probabilities.

\subsection{Results on Translated English Test Sets}
\label{sec:appendix-result-on-english}

To ensure translation quality, we used the gpt-4-1106-preview model. 
The translation examples listed in Table~\ref{tbl:translation} demonstrate that this model can accurately translate texts containing categories closely related to Chinese culture.

\begin{CJK}{UTF8}{gbsn}
\begin{table}[ht]
	\centering
	\small
	\begin{tabular}{L{0.21\textwidth}|L{0.21\textwidth}}
		\hline
		\bf Chinese & \bf Translated English  \\
		\hline
		穿蓝色上衣的男人拿着的唢呐 & The suona held by the man in the blue shirt \\
        \hline
        距离岸堤最近的一艘龙舟 & The dragon boat closest to the shore \\
        \hline
        穿着深色唐装的人在拉的二胡 & The erhu being played by the person in the dark Tang suit \\
        \hline
        头扭向一侧且戴着眼镜的女生左手拿的琵琶 & The pipa held in the left hand of the female with her head turned to one side and wearing glasses \\
        \hline
        里面放有许多绿色青菜的那个火锅 & The hotpot containing many green vegetables \\
		\hline
	\end{tabular}
	\caption{Examples of original Chinese and translated English in the CVLUE VG test set.}
	\label{tbl:translation}
\end{table}
\end{CJK}

\begin{figure}[htbp]
  \centering
  \includegraphics[width=0.48\textwidth]{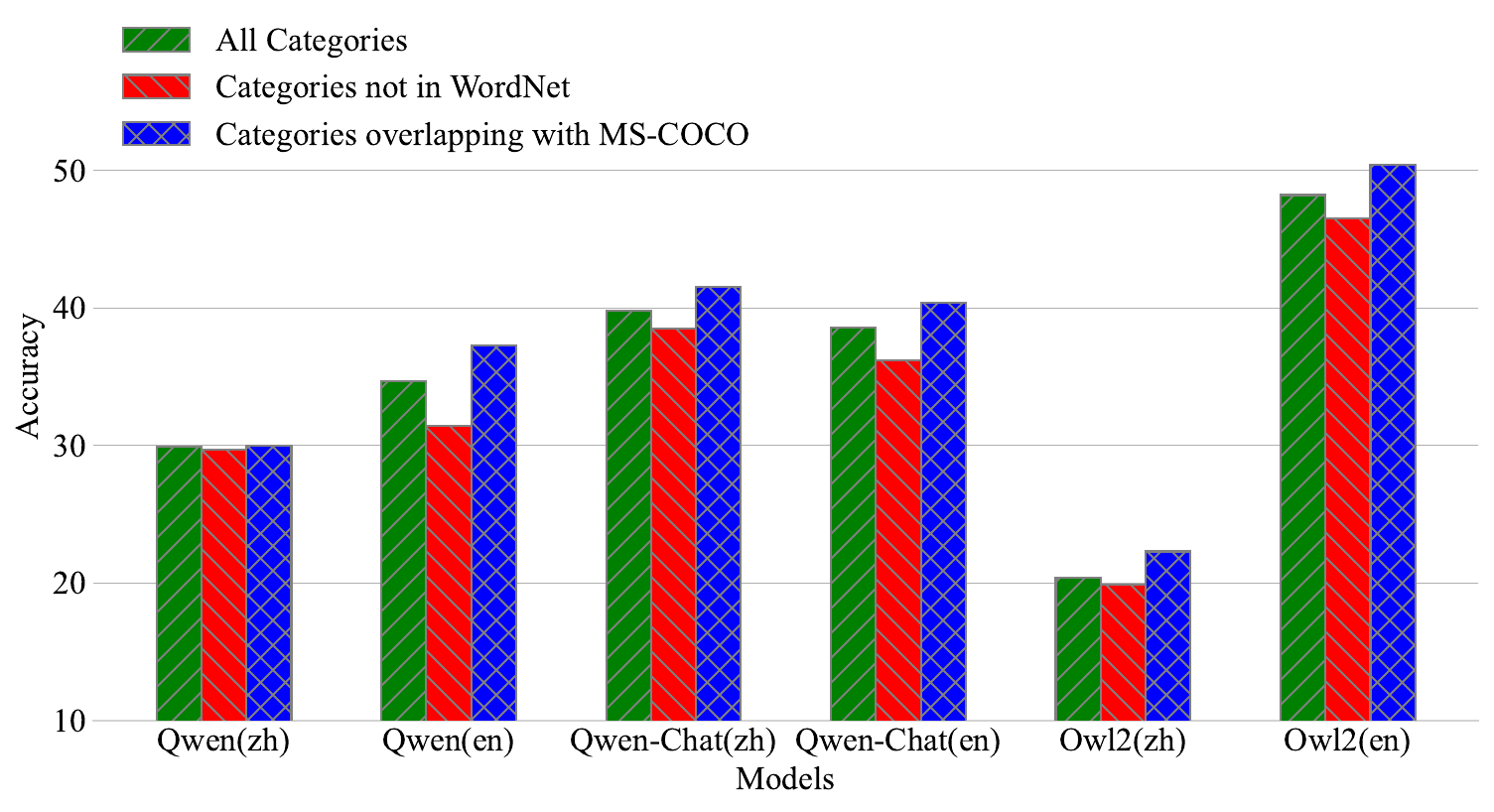}
  \caption{Category group results of QwenVL, QwenVL-Chat and mPLUG-Owl2 on the original Chinese (zh) and translated English (en) CVLUE VQA test set.}
  \label{fig:vqa_en_vs_zh_by_group}
\end{figure}

Figure~\ref{fig:vqa_en_vs_zh_by_group} shows the results of QwenVL, QwenVL-Chat, and mPLUG-Owl2 on the original Chinese test set and the translated English test set for the CVLUE VQA task. 
It is worth noting that the mPLUG-Owl2 model exhibits a significant performance gap between the original Chinese and translated English test sets. 
Analysis of the prediction results reveals that this discrepancy is due to the model's misunderstanding of the prompt. 
Despite the input prompt explicitly instructing the model to answer in Chinese (see Appendix~\ref{sec:appendix-vqa-prompt}), 56\% of the model's predictions were still in English. 
Therefore, the model's performance on the Chinese test set does not fully reflect its knowledge of Chinese culture.

\subsection{Zero-Shot vs. Fine-Tuninig}
\label{sec:appendix-zs-vs-ft}

\begin{figure}[htbp]
  \centering
  \includegraphics[width=0.48\textwidth]{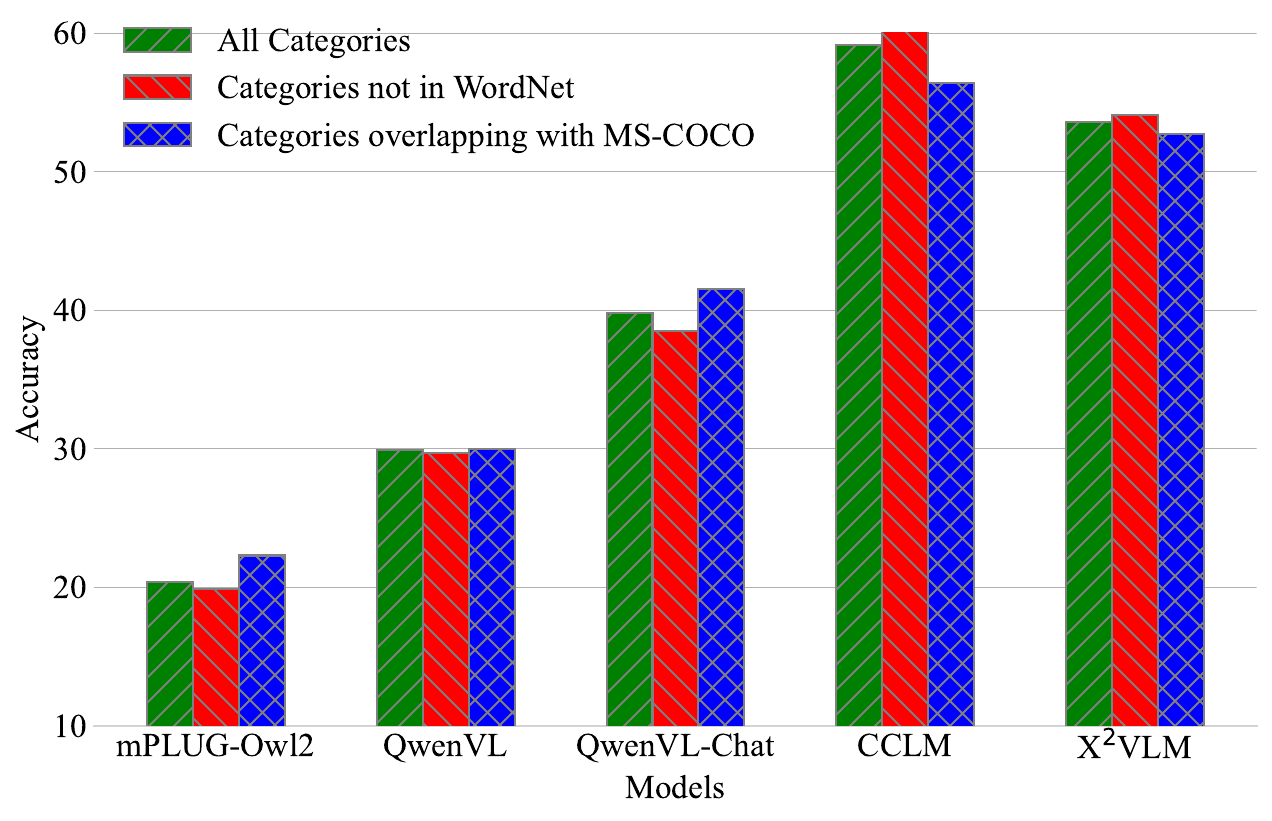}
  \caption{Category group results on CVLUE VQA task.}
  \label{fig:vqa_avg_by_group}
\end{figure}

The category group results of zero-shot and fine-tuned models on VQA are shown in Figure~\ref{fig:vqa_avg_by_group}.

\subsection{Results by Category}
\label{sec:appendix-result-by-cat}

\begin{figure*}[htbp]
  \centering
  \includegraphics[width=0.95\textwidth]{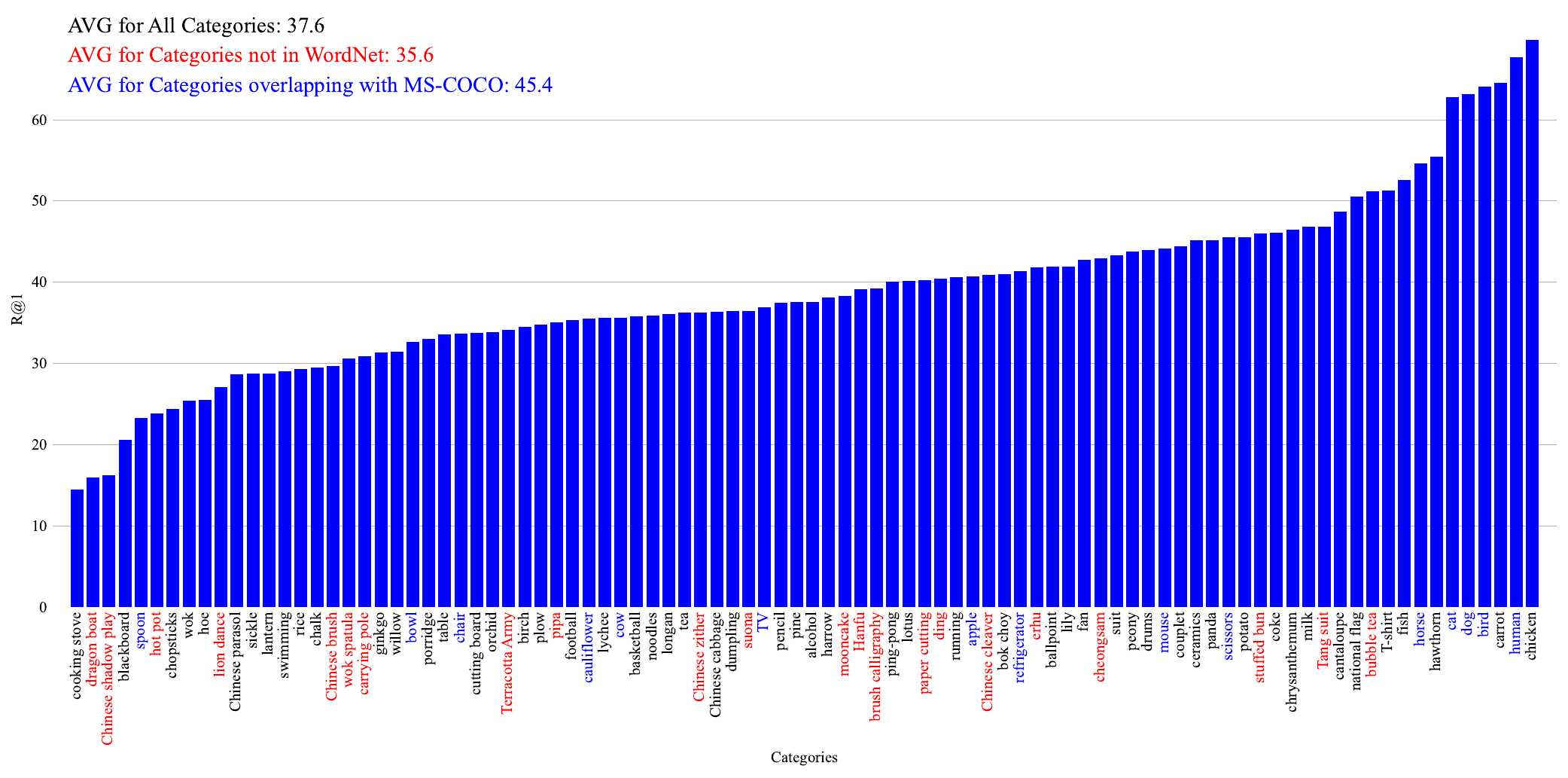}
  \caption{Results of X$^2$VLM model on the CVLUE IR task, displayed by image category.}
  \label{fig:ir_x2vlm_by_cat}
\end{figure*}

\begin{figure*}[htbp]
  \centering
  \includegraphics[width=0.95\textwidth]{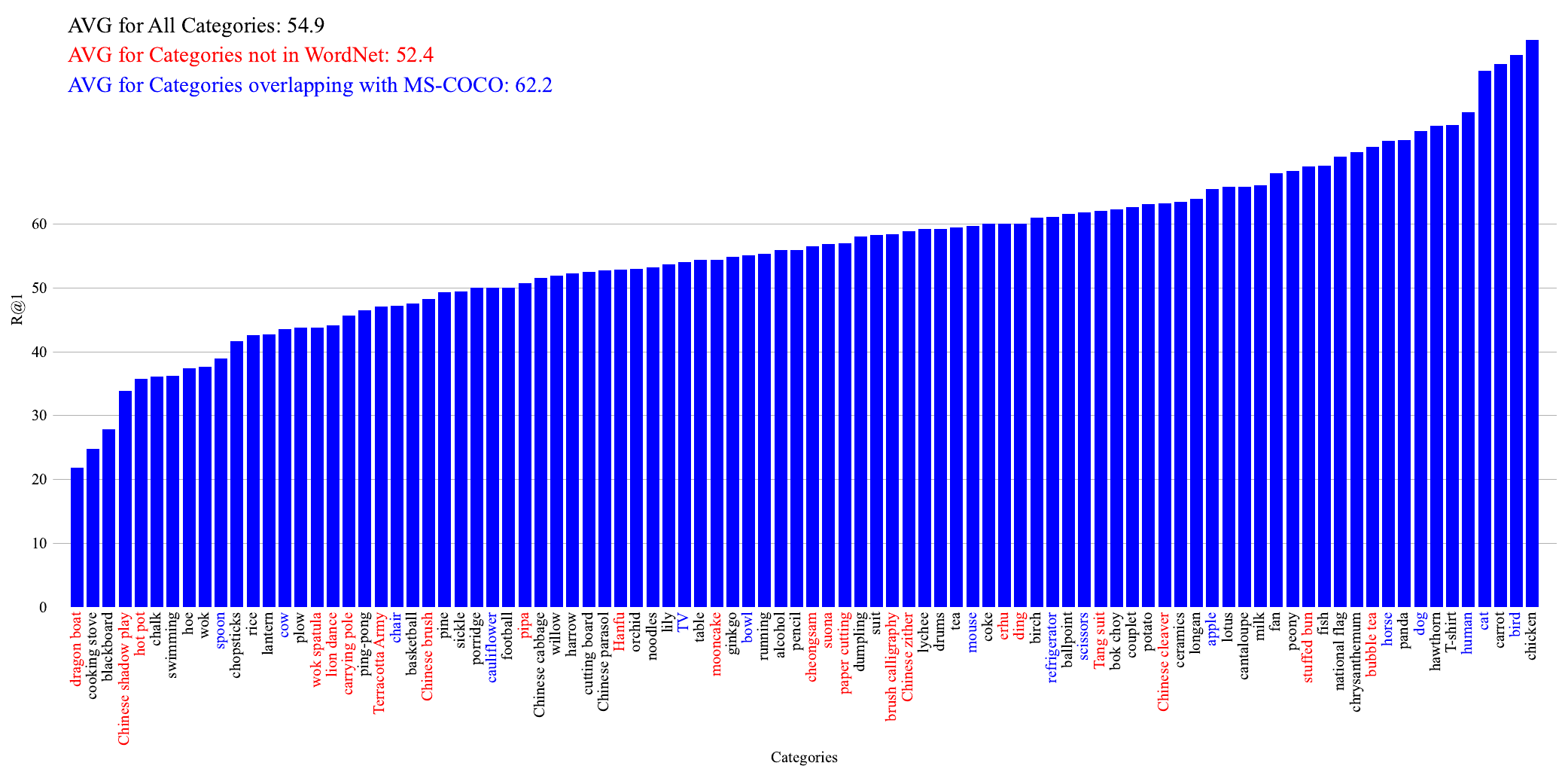}
  \caption{Results of X$^2$VLM model on the CVLUE TR task, displayed by image category.}
  \label{fig:tr_x2vlm_by_cat}
\end{figure*}

\begin{figure*}[htbp]
  \centering
  \includegraphics[width=0.95\textwidth]{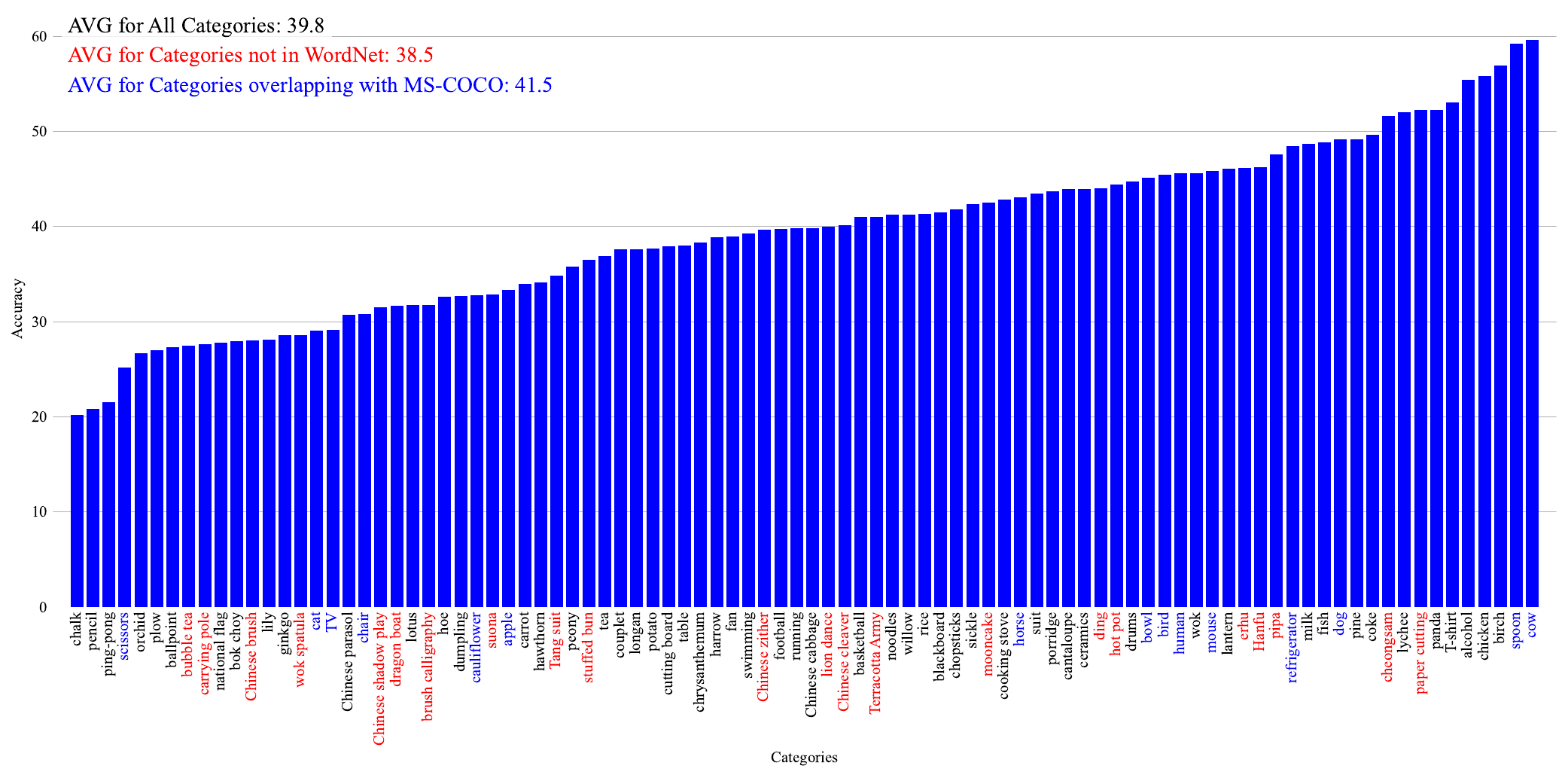}
  \caption{Results of QwenVL-Chat model on the CVLUE VQA task, displayed by image category.}
  \label{fig:vqa_qwen_chat_by_cat}
\end{figure*}

\begin{figure*}[htbp]
  \centering
  \includegraphics[width=0.95\textwidth]{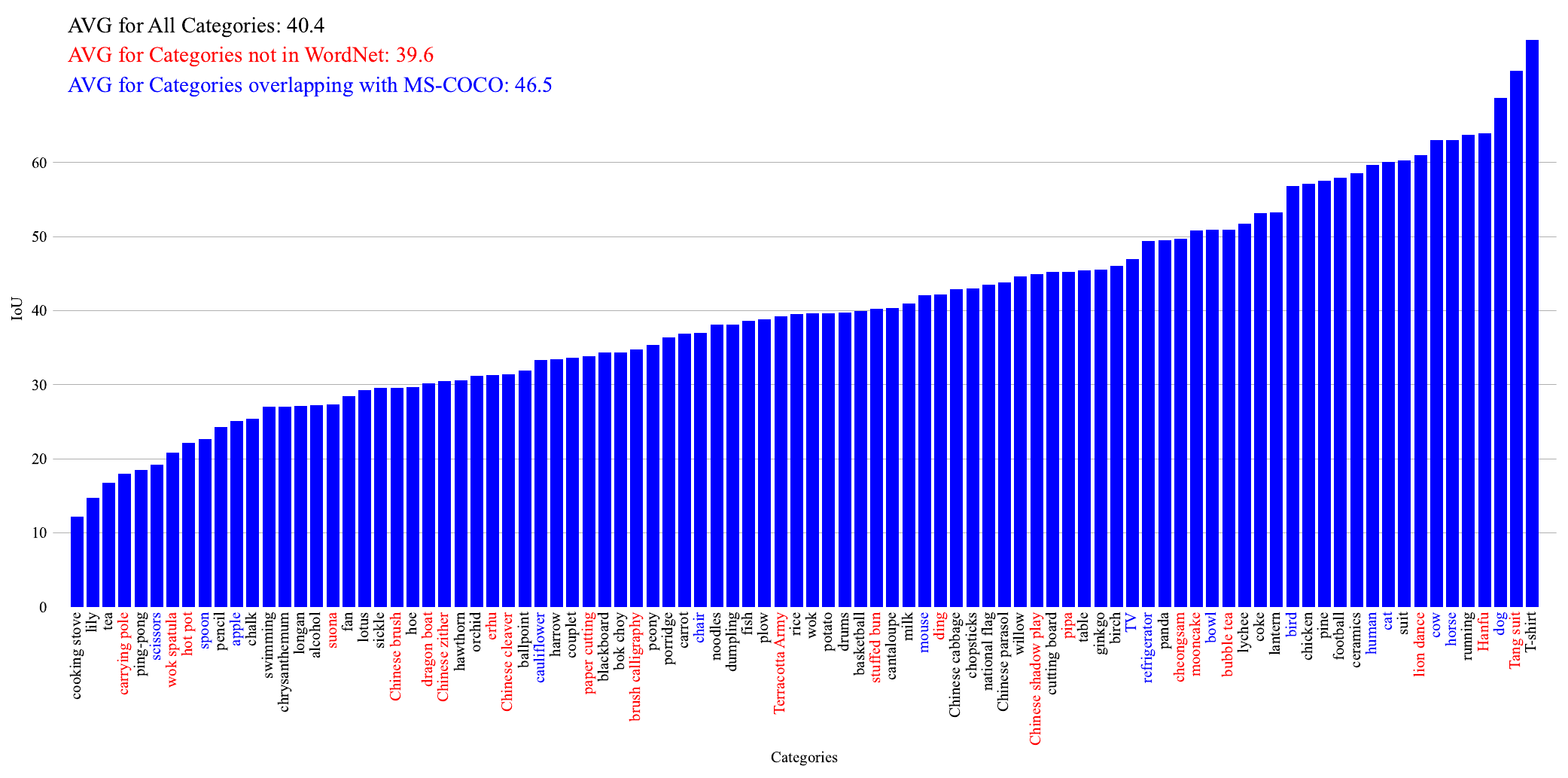}
  \caption{Results of QwenVL-Chat model on the CVLUE VG task, displayed by image category.}
  \label{fig:vg_qwen_chat_by_cat}
\end{figure*}

\begin{figure*}[htbp]
  \centering
  \includegraphics[width=0.95\textwidth]{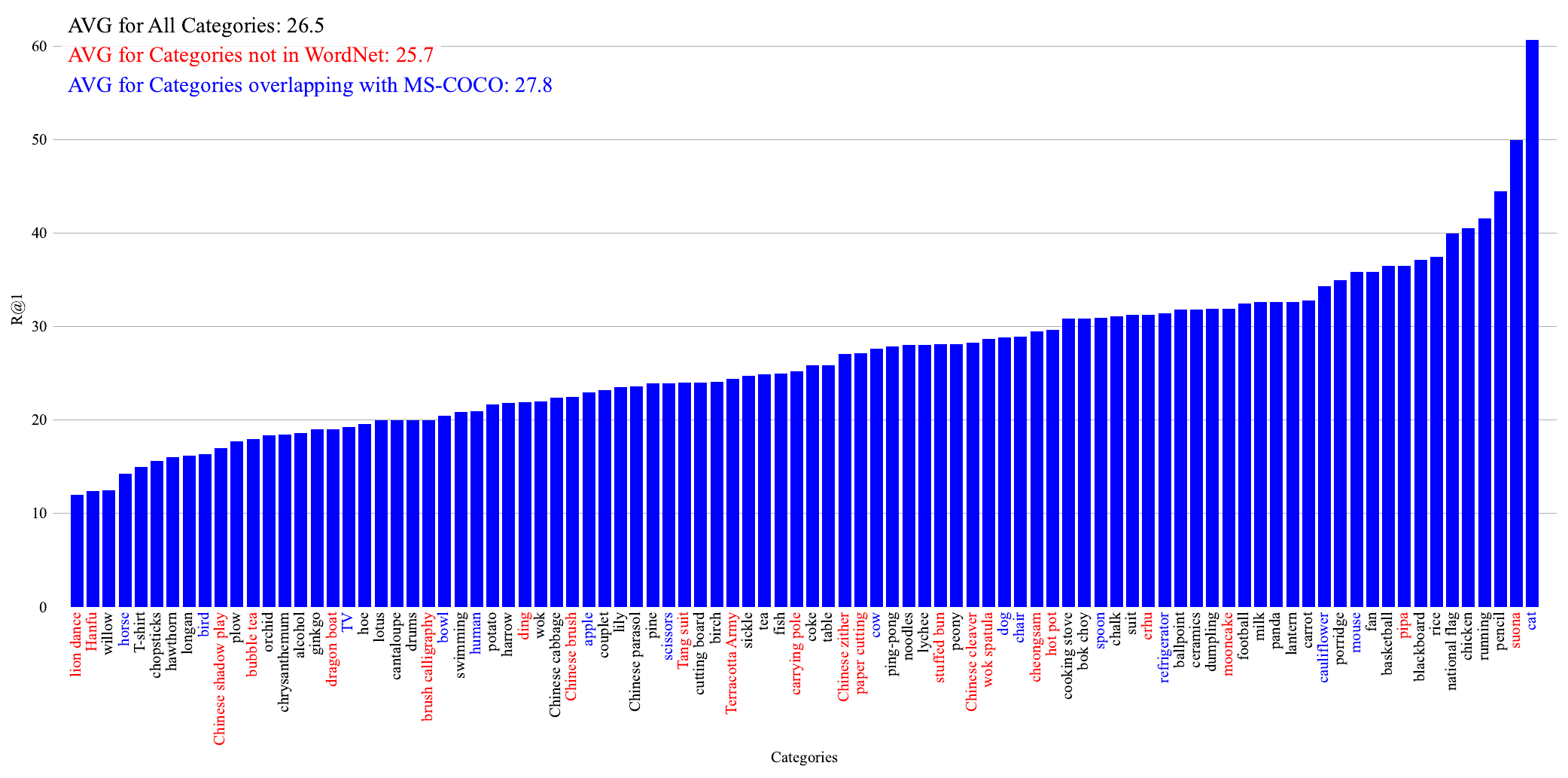}
  \caption{Results of QwenVL-Chat model on the CVLUE VD task, displayed by image category.}
  \label{fig:vd_qwen_chat_by_cat}
\end{figure*}

The results by category on IR, TR, VQA, VG and VD tasks are shown in Figure~\ref{fig:ir_x2vlm_by_cat}, \ref{fig:tr_x2vlm_by_cat}, \ref{fig:vqa_qwen_chat_by_cat}, \ref{fig:vg_qwen_chat_by_cat} and \ref{fig:vd_qwen_chat_by_cat}, respectively.

\end{document}